\documentclass{article}
\usepackage{iclr2027_conference,times}

\usepackage{hyperref}
\usepackage{url}
\usepackage{graphicx}
\usepackage{amsmath}
\usepackage{amsthm}
\usepackage{amssymb}
\usepackage{booktabs}
\usepackage{enumitem}
\usepackage{multirow}
\usepackage{algorithm}
\usepackage{algorithmicx}
\usepackage{algpseudocode}
\usepackage{xcolor}
\usepackage{tabularx}
\usepackage{subcaption}

\DeclareMathOperator*{\argmin}{arg\,min}

\newcommand{\model}{EpiMind}

\title{Graph World Models for Constrained \\Epidemic Policy Planning}

\author{
Yiqi Su$^{1}$ \quad
Rashed Shelim$^{1,2}$ \quad
Lingyi Wang$^{2}$ \quad
Walid Saad$^{2}$ \quad
Naren Ramakrishnan$^{1}$ \\[3pt]
$^{1}$Department of Computer Science, Virginia Tech,
Alexandria, VA 22305, USA \\
$^{2}$Department of Electrical and Computer Engineering, Virginia Tech,
Alexandria, VA 22305, USA
}

\iclrfinalcopy

\makeatletter
\patchcmd{\@maketitle}
  {\lhead{Published as a conference paper at ICLR 2027}}
  {\lhead{}}
  {}
  {\PackageError{arxiv}{Could not replace the ICLR header}{}}
\makeatother



\begin{document}

\maketitle

\begin{abstract}
Epidemic
policy planning
often requires
coordination
between geographical regions, taking into account mobility-driven spillovers and how to make use of limited resources.
Existing methods either lack action-conditioned models of coupled dynamics or cannot guarantee per-period feasibility.
We present \model{}, a graph world model framework for constrained epidemic policy planning across regions. A graph-factored recurrent state-space model generates joint policy-conditioned rollouts from regional latent beliefs, while graph-temporal ADMM optimizes regional interventions, enforces shared-resource feasibility through projection, and evaluates temporal specifications under the learned model.
EpiMind reduces admission RMSE by 29\% relative to graph-free dynamics modeling, plans within 1--5\% of the best feasible constant policy with guaranteed shared-budget feasibility, and outperforms all deployable baselines across three resource budgets in real-context evaluation.
These results demonstrate that graph-structured policy imagination with explicit constrained coordination supports effective epidemic interventions from learned dynamics.
\end{abstract}

\section{Introduction}
The effective control of epidemics requires coordinating non-pharmaceutical, pharmaceutical, and surveillance interventions across multiple regions~\cite{kraemer2020mobility,ferretti2020quantifying,hsiang2020effect}. The COVID-19 epidemic demonstrated how a patchwork of responses created uncertainty, chaos, and ultimately lack of trust in public health authorities~\cite{birkland2021governing,steelfisher2023trust}. Jurisdictions act on observable local conditions but remain coupled through mobility and competition for finite vaccines, hospital capacity, and budgets~\cite{emanuel2020fair}.

Coordinating policy making across regions is difficult, especially under uncertainty and shared resource constraints~\cite{kermack1927sir,balcan2010glam}.
A given surveillance trend may reflect transmission changes, altered testing, or voluntary behavioral adaptation across regions. A useful planning model must therefore infer latent epidemic conditions from noisy, delayed, policy-dependent observations and predict their graph-coupled evolution under alternative actions. 

Existing methods address only a fragment of this problem. Compartmental models~\cite{kermack1927sir} and their metapopulation extensions~\cite{arino2003metapopulation,balcan2010glam} simulate forward trajectories by modifying mechanistic parameters such as transmission rates or contact matrices. They
treat policies as exogenous inputs rather than decision variables.
Further, they assume reported cases are direct measurements of true incidence rather than policy-dependent surveillance signals. Graph neural network (GNN)-based spatiotemporal forecasters such as Cola-GNN~\cite{deng2020colagnn} and county-level COVID-19 predictors~\cite{kapoor2020examining} learn flexible dynamics from time series but have no action space; thus, they cannot distinguish whether a forecasted decline reflects an intervention's effect or a confounder such as reduced testing. Reinforcement learning (RL) for epidemic control~\cite{kompella2020epirl,ohi2020exploring,bushaj2023vaccine} reframes the problem as sequential decision-making within a calibrated simulator but typically optimizes a single composite agent over a concatenated multi-region state, with resource limits absorbed into shaped rewards that provide no feasibility guarantee for hard joint constraints such as total vaccine supply summed across regions. Multi-agent RL approaches that extend MAPPO~\cite{yu2022mappo} to regional epidemic control~\cite{nayak2023epimarl} train per-region policies under a shared critic, but cross-agent sum constraints are typically incorporated as reward penalties or adaptive Lagrange multipliers, which guarantee only expected feasibility at convergence rather than per-timestep feasibility for a planner facing fixed inventory, and require a high-fidelity simulator that is unavailable for active outbreak response. Mathematical programming methods such as mixed-integer optimization for vaccination facility location~\cite{bertsimas2022vaccine} and vaccine supply-chain optimization~\cite{duijzer2018vaccine} enforce hard constraints exactly through branch-and-cut, but require the transmission dynamics (case trajectory, susceptible fraction, reproduction number) to be supplied as a pre-fit input from a separately calibrated SEIR model, so the optimizer cannot adapt as decisions, behavioral response, or variant emergence shifts the trajectory.

World models~\cite{hafner2023dreamerv3,wang2025dmwm,memon2026toward} address several of these limitations by learning latent transition and observation models that can be rolled forward under candidate action sequences without further interaction with the environment. Conditioning the dynamics on actions allows the model to represent policy-dependent evolution and surveillance. It does not, however, identify causal intervention effects under endogenous historical policies, where regions may adopt stronger interventions precisely when outbreaks worsen. Graph world models (GWMs)~\cite{feng2025graph} extend recurrent state-space models (RSSMs)~\cite{hafner2023dreamerv3} by representing interacting entities as nodes that exchange information through message passing, providing an appropriate inductive bias for mobility-coupled epidemics. Existing world-model methods nevertheless provide limited machinery for coordinating distinct regional actions under per-period shared-resource constraints,
and this is a gap we propose to address.
We introduce \model{}, a graph world-model planning framework 
whose key contributions are:

\begin{itemize}
    \item
    \model{} helps formulate multi-region epidemic planning on a dynamic policy graph, coupling 
    regional dynamics through mobility and finite shared resources.

    \item
    \model{} couples a parameter-shared GWM with a constrained multi-region planner for joint policy rollout, resource-feasible allocation, and temporal-logic evaluation.

    \item
    \model{} achieves near-oracle performance in real-context evaluation, reduces admissions by 53.4\% relative to no intervention while satisfying all shared-resource constraints.

\end{itemize}

\section{Method}

\subsection{Problem Formulation}
\label{sec:problem_form}

\paragraph{Epidemics on graphs.}
\label{subsec:graph}


We formalize the regional epidemic by a time-varying policy graph $G_t=(V_t,E_t)$ (see details in Section~\ref{sec:epi_foundations}), where nodes $V_t$ represent regions and edges $E_t$ capture mobility and coordination constraints at time $t$. This formulation makes the planning problem tractable by aligning each computational ingredient with the epidemic's physical structure. 

\paragraph{Graph-structured latent dynamics.}
\label{subsec:state_action_dynamics}

Let $x_t^i$ denote region $i$'s unobserved epidemic state, and $\mathbf x_t=(x_t^1,\ldots,x_t^N)$ denote the joint state. Each region selects a $D$-dimensional intervention vector:
\begin{equation}
    a_t^i \in \mathcal A^i \subseteq [0,1]^D,
    \qquad
    \mathbf a_t=(a_t^1,\ldots,a_t^N),
    \label{eq:joint_action}
\end{equation}
The action comprises local non-pharmaceutical intervention (NPI) intensity and regional allocations of shared vaccine, hospitalization capacity, and fiscal-resource budget.
The joint transition is policy-conditioned and graph-coupled:
\begin{equation}
p_\theta(\mathbf x_{t+1}\mid \mathbf x_t,\mathbf a_t,G_t)
=
\prod_{i=1}^{N}
p_\theta\!\left(
x_{t+1}^i
\,\middle|\,
x_t^i,a_t^i,
\operatorname{Agg}_{j\in\mathcal N_t(i)}
\!\left(E_t^{ij},x_t^j,a_t^j\right)
\right).
\label{eq:graph_dynamics}
\end{equation}
Here \(\mathcal N_t(i)\) is region \(i\)'s mobility neighborhood, and \(\theta\) is shared across regions. Regional states, actions, observations, and neighborhoods remain distinct.
The latent state is not observed directly. Instead, region $i$ receives a surveillance observation, e.g., infections, hospital admissions, or deaths. 
\begin{equation}
    o_t^i
    \sim
    \Omega_\theta^i
    \!\left(
        \cdot \mid x_t^i,\mathbf a_{t-1}
    \right),
    \label{eq:observation_ma}
\end{equation}
Candidate policies are evaluated over a multi-step horizon rather than through one-step prediction alone to response to the delayed surveillance and intervention effects.

\paragraph{Constrained policy planning.}
\label{subsec:planning_problem}

At decision epoch $t$, the planner evaluates a candidate sequence of joint regional actions $\mathbf a_{t:t+H-1}$ over horizon $H$. Starting from the current regional
beliefs, a learned dynamics model $M_\theta$ generates the joint policy-conditioned rollout:
\begin{equation}
    \hat{\tau}_{t+1:t+H}^{1:N}
    =
    M_\theta
    \!\left(
        b_t^{1:N},z_t^{1:N},
        \mathbf a_{t:t+H-1},G_t
    \right).
    \label{eq:joint_rollout}
\end{equation}
These rollouts support model-dependent comparisons among candidate policies and, in simulation, are validated against known counterfactual outcomes. The planner minimizes predicted health and intervention costs subject to shared resource budgets:

\begin{equation}
\begin{aligned}
\min_{\mathbf a_{t:t+H-1}}
\quad&
\sum_{\tau=t}^{t+H-1}
\ell\!\left(
    \hat{\tau}_{\tau}^{1:N},
    \mathbf a_\tau
\right)
-
\beta\,
\widetilde{\rho}_{\varphi}
\!\left(
    \hat{\tau}_{t+1:t+H}^{1:N}
\right)
\\
\text{s.t.}\quad&
\sum_{i=1}^{N} a_{\tau}^{i,d}
\le B_{\tau}^{d},
\qquad
d\in\mathcal D_{\mathrm{shared}},
\\
&
0\le a_{\tau}^{i,d}\le1,
\qquad
i=1,\ldots,N,\quad
\tau=t,\ldots,t+H-1,
\end{aligned}
\label{eq:planning_problem}
\end{equation}

where \(\ell\) balances predicted epidemic burden and intervention cost, \(\widetilde{\rho}_{\varphi}\) is a differentiable robustness score for temporal specification \(\varphi\), and \(\mathcal D_{\mathrm{shared}}\) indexes the resource-constrained action dimensions. 
Projection guarantees that the executed action satisfies the specified resource budgets, and temporal constraints are evaluated relative to the learned model by rerolling the projected action.

\subsection{\model\ Framework}
\label{sec:framework}

\begin{figure}[t]
    \centering
    \includegraphics[width=\linewidth]
    {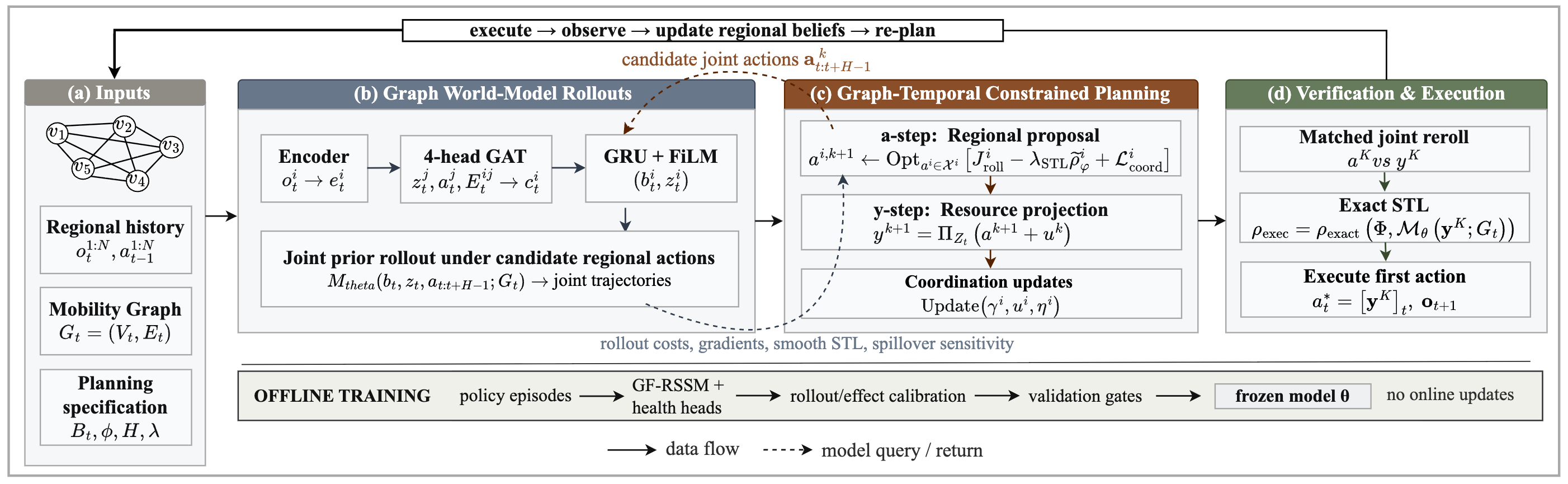}
    \vspace{-7mm}
    \caption{\model{} framework. A parameter-shared GF-RSSM updates regional beliefs and generates graph-coupled policy rollouts. GT-ADMM coordinates and projects regional actions, rerolls the feasible allocation for model-relative STL evaluation, and executes its first action.}
    \label{fig:pipeline}
    \vspace{-3mm}
\end{figure}

\model{} couples a learned GWM with constrained receding-horizon planning as shown in Figure~\ref{fig:pipeline}. The world model predicts joint epidemic trajectories under candidate regional interventions, and the planner coordinates the interventions subject to the shared resource constraints. 

\subsubsection{Graph World-Model Rollouts}
\label{subsec:world_model_planning}


\paragraph{Graph-factored recurrent state-space model (GF-RSSM).}
We instantiate the dynamics in Section~\ref{sec:problem_form} with a graph-factored recurrent state-space model (GF-RSSM). Neural parameters are shared across regions, while each region \(i\) maintains its own recurrent belief \(b_t^i\) and stochastic state \(z_t^i\). Neighboring states and actions are aggregated through graph attention:

\begin{align}
c_t^i
&=
\operatorname{GAT}_{\theta}
\!\left(
 \{z_t^j,a_t^j,E_t^{ij}\}_{j\in\mathcal N(i)}
\right),
\label{eq:gat_context}\\
b_t^i
&=
f_{\theta}
\!\left(
b_{t-1}^i,z_{t-1}^i,a_{t-1}^i,c_{t-1}^i
\right),
\label{eq:gf_rssm_det}\\
z_t^i
&\sim
q_{\theta}
\!\left(
z_t^i\mid b_t^i,e_{\theta}(o_t^i)
\right),
\qquad
\hat o_t^i
\sim
p_{\theta}
\!\left(
o_t^i\mid b_t^i,z_t^i
\right).
\label{eq:gf_rssm_obs}
\end{align}

\paragraph{Joint policy-conditioned rollout.}
The graph context \(c_t^i\) transmits mobility-weighted information from neighboring regions. 
Parameter sharing provides a common transition model without imposing identical regional trajectories: beliefs, latent states, observations, actions, and neighborhoods remain region specific.
The posterior in~\eqref{eq:gf_rssm_obs} assimilates the current observation. Future observations are unavailable during planning, so imagined trajectories use the learned prior recursively. At every rollout step, all regional states advance jointly under the complete action vector and mobility graph:
\begin{equation}
\hat{\mathbf x}_{t+1:t+H}
=
\mathcal M_{\theta}
\!\left(
\mathbf b_t,\mathbf z_t,
\mathbf a_{t:t+H-1};
G_{t:t+H-1}
\right).
\label{eq:joint_rollout}
\end{equation}

Thus, each trajectory depends on both local and neighboring interventions. Dedicated admission and occupancy heads decode the health quantities used by the planner.

\paragraph{World-model training.}
The GF-RSSM is trained offline for predictive and policy-effect fidelity:
\begin{equation}
\mathcal L_{\mathrm{WM}}
=
\mathcal L_{\mathrm{obs}}
+\mathcal L_{\mathrm{reward}}
+\mathcal L_{\mathrm{KL}}
+\mathcal L_{\mathrm{rollout}}
+\mathcal L_{\mathrm{effect}}
+\mathcal L_{\mathrm{cal}},
\label{eq:s1_loss}
\end{equation}
where the first three terms form the recurrent state-space objective; the rollout, effect, and calibration terms supervise delayed policy responses. The trained world model is frozen during planning.

\subsubsection{Graph-Temporal Constrained Planning}
\label{subsec:gt_admm}

\paragraph{Graph-temporal alternating direction method of multipliers (GT-ADMM).}
We implement the constrained planner as graph-temporal alternating direction method of multipliers (GT-ADMM), which alternates regional action optimization, shared-resource projection, and coordination updates.
At decision epoch \(t\), GT-ADMM optimizes region-specific action sequences
over horizon \(H\):
\begin{equation}
\begin{aligned}
\min_{\mathbf a}\quad
J_{\theta}(\mathbf a)
&=
\sum_{i=1}^{N}
f^i\!\left(
\hat{\mathbf x}_{t+1:t+H},\mathbf a^i
\right)
-
\lambda_{\mathrm{STL}}
\sum_{i=1}^{N}
\rho_{\mathrm{sm}}\!\left(
\Phi^i,\hat{\mathbf x}^{\,i}_{t+1:t+H}
\right), \\
\hat{\mathbf x}_{t+1:t+H}
&=
\mathcal M_{\theta}(\mathbf a;G_t),
\end{aligned}
\label{eq:planning_objective}
\end{equation}
subject to local action bounds, edge relations, and the shared resource set \(\mathcal Z_t\). Here \(f^i\) is the per-region health--intervention cost:
\begin{equation}
f^i
=
\sum_{h=1}^{H}
\left[
\ell_{\mathrm{health},t+h}^i
+
\lambda_{\mathrm{NPI}}
\ell_{\mathrm{NPI},t+h}^i
\right],
\label{eq:regional_rollout_cost}
\end{equation}
where \(\lambda_{\mathrm{NPI}}\) controls the trade-off between predicted
health burden and NPI burden. The coefficient
\(\lambda_{\mathrm{STL}}\) weights the smooth robustness
\(\rho_{\mathrm{sm}}\) of temporal specification \(\Phi^i\).
GT-ADMM alternates regional proposal updates, shared-resource projection,
and coordination updates.

\paragraph{Regional proposal update.}
Each region updates its action block while holding the other regions at their latest reference actions:
\begin{equation}
\begin{aligned}
a^{i,k+1}
=\argmin_{a^i\in\mathcal X^i}\Bigg[
&\mathcal J_{\mathrm{roll}}^i
+\frac{\rho_e}{2}\sum_{j\in\mathcal N(i)}
 \lVert a^i-a^{j,k}\rVert^2
+(\gamma^{i,k})^\top a^i \\
&+\frac{\rho_g}{2}
 \lVert a^i-y^{i,k}+u^{i,k}\rVert^2
+(\eta^{i,k})^\top s^i
+\frac{\sigma}{2}\lVert a^i-a^{i,k}\rVert^2
\Bigg],
\end{aligned}
\label{eq:x_step}
\end{equation}
where
\begin{equation}
\mathcal J_{\mathrm{roll}}^i
=
f^i\!\left(a^i,\mathcal M_\theta
(a^i,\mathbf a^{-i,k};G_t)\right)
-
\beta\,\rho_{\mathrm{sm}}
\!\left(\Phi^i,\mathcal M_\theta
(a^i,\mathbf a^{-i,k};G_t)\right).
\end{equation}
The rollout remains joint so that changing \(a^i\) can alter the predicted outcomes of every connected region. The remaining terms encourage neighboring-policy agreement, consistency with the feasible allocation, spillover awareness, and stable successive updates. Equation~\eqref{eq:x_step} is solved by gradient descent through the frozen world model.
The spillover signal is
\begin{equation}
s^i
=
\sum_{j\in\mathcal N(i)}
E_t^{ij}
\nabla_{a^i}
g_{\mathrm{spill}}^j(x_t,a^i),
\label{eq:spillover_signal}
\end{equation}
where \(g_{\mathrm{spill}}\) predicts neighboring outcome changes. This signal is predictive, not causally identified.

\paragraph{Resource projection.}
Regional proposals need not be jointly feasible. GT-ADMM therefore computes the nearest allocation in the shared resource set:
\begin{equation}
y^{k+1}
=
\argmin_{z\in\mathcal Z_t}
\sum_{i=1}^{N}
\frac{\rho_g}{2}
\left\lVert
a^{i,k+1}-z^i+u^{i,k}
\right\rVert^2 .
\label{eq:z_step}
\end{equation}
Because \(\mathcal Z_t\) contains linear box and sum constraints, the \(z\)-step decomposes by resource into capped-simplex projections. The planner executes \(y^{k+1}\), guaranteeing satisfaction of these budgets.

\paragraph{Coordination updates.}
After projection, the coordination variables are updated as
\begin{align}
\gamma^{i,k+1}
&=
\gamma^{i,k}
+
\rho_e
\sum_{j\in\mathcal N(i)}
\left(a^{i,k+1}-a^{j,k+1}\right),
\label{eq:edge_dual}\\
u^{i,k+1}
&=
u^{i,k}
+
a^{i,k+1}-y^{i,k+1},
\label{eq:global_dual}\\
\eta^{i,k+1}
&=
\eta^{i,k}
+
\alpha s^i,
\label{eq:spillover_update}
\end{align}
where \(u^i\) reflects pressure from shared-resource scarcity, \(\gamma^i\) tracks disagreement with mobility-connected regions, and \(\eta^i\) tracks predicted cross-region spillover sensitivity.  
The node-level \(\gamma^i\) and \(\eta^i\) are approximate accumulators for neighbor disagreement and predicted spillover sensitivity, respectively. These approximations provide interpretable coordination signals.

\subsubsection{Projection, Verification, and Execution}
\label{subsec:verification}

After the final iteration, \model{} retains the unprojected proposal \(a^K\) and executes the projected allocation \(y^K\). Both are rerolled through the same joint world-model interface. Smooth STL robustness is used to obtain gradients during optimization, whereas exact nonsmooth robustness is evaluated on the projected trajectory:
\begin{equation}
\rho_{\mathrm{exec}}
=
\rho_{\mathrm{exact}}
\!\left(
\Phi,
\mathcal M_\theta(y^K;G_t)
\right).
\label{eq:executed_stl}
\end{equation}
Resource feasibility is guaranteed for the constraints represented in \(\mathcal Z_t\). In contrast, STL satisfaction is model relative, i.e., a positive \(\rho_{\mathrm{exec}}\) certifies the learned rollout, not the unknown true environment. 
The first action of \(y^K\) is executed, and the resulting joint observation updates the regional posterior at decision epoch \(t+1\). This produces the closed-loop sequence:
\textit{$\text{belief update}
\;\rightarrow\;
\text{joint rollout}
\;\rightarrow\;
\text{constrained planning}
\;\rightarrow\;
\text{projection and verification}
\;\rightarrow\;
\text{execution}$}.

\section{Experiments}
\label{sec:experiments} 


We evaluate \model{} through five research questions spanning predictive fidelity, planning utility, constraint handling, coordination, and real-context transfer: 

\begin{itemize}
\item[\textbf{RQ1}] \textbf{Predictive fidelity.}
Does the graph world model accurately predict held-out trajectories and policy responses? (Section~\ref{subsec:world_model_results},
Appendix~\ref{app:wm_extended}) 

\item[\textbf{RQ2}] \textbf{Planning effectiveness.}
Does planning through learned joint rollouts improve the matched health--intervention objective? (Section~\ref{subsec:planning_results}, Appendix~\ref{app:robustness})

\item[\textbf{RQ3}] \textbf{Feasibility and verification.}
Does projection enforce shared budgets, and do projected rollouts satisfy model-relative temporal specifications? (Section~\ref{subsec:constraint_results},
Appendix~\ref{app:planning_extended})

\item[\textbf{RQ4}] \textbf{Graph coordination.}
Does mobility-aware coordination improve allocation under heterogeneous conditions and shared resources (Section~\ref{subsec:coordination_results},
Appendix~\ref{app:coord_ablation})

\item[\textbf{RQ5}] \textbf{Real-context transfer.}
Does \model{} support forecasting and allocation in realistic multi-region settings? (Section~\ref{subsec:real_data},
Appendices~\ref{app:mobility}--\ref{app:realcontext_extended})

\end{itemize}

\subsection{Experimental Protocol} 
\label{subsec:settings} 

\paragraph{Evaluation tracks.} 
We evaluate \model{} in three complementary settings: (1) a mobility-coupled multi-region simulator provides known dynamics and ground-truth outcomes for evaluating policy-conditioned prediction and planning; (2) a retrospective U.S.\ state-level panel evaluates forecasting and allocation behavior under observed surveillance, intervention, capacity, and mobility data; and (3) a real-context semi-simulated setting which initializes the simulator from real data while retaining known dynamics for realized evaluation of alternative policies.

\paragraph{Datasets.} The synthetic benchmark contains \(N{=}5\) mobility-coupled regions over \(T{=}26\) weekly decision epochs. Regional actions control NPI intensity and allocations of vaccine, hospitalization-capacity, and fiscal resources under shared budgets. The planner observes delayed, noisy surveillance signals rather than latent SEIR states; Table~\ref{tab:params} reports the complete simulator configuration. 

For the real-context evaluation, we construct a weekly U.S.\ state-level panel combining reported cases, deaths, hospital admissions and capacity, vaccination, policy interventions, population, and directed interstate mobility. Track~A retrospectively evaluates forecasting and model-relative policy projections at held-out decision origins; outcomes under unexecuted policies are unavailable. Track~B initializes a semi-synthetic simulator from the same regional conditions and mobility graph, permitting realized evaluation of alternative policies under known dynamics. Data sources and preprocessing appear in Appendix~\ref{app:real_data}.

    \vspace{-3mm}

\paragraph{Benchmarking methods.} World-model comparisons include statistical predictors, action-conditioned sequence models, and graph ablations. Planning comparisons include constant and heuristic policies, MPC, ADMM, RL-based controllers, graph-free and independent variants, and an oracle-dynamics reference. All policies are evaluated under the same action bounds, resource budgets, projection, and health--intervention objective. Implementation details appear in Appendix~\ref{app:baselines}.



\paragraph{Metrics.}
For RQ1, we report held-out admission MAE, RMSE, cumulative rollout error, and policy-conditioned dose response. For RQ2, we report admissions per 100K, NPI burden, matched objective \(J\), and regret relative to the best feasible constant policy. For RQ3, we report budget feasibility, maximum excess, projection displacement, and exact model-relative STL robustness. For RQ4, we compare realized objective values and paired outcomes under shared budgets, supplemented by a stepwise matched-burden ablation. For RQ5, we report retrospective model-relative comparisons and realized outcomes in the real-context benchmark.

\subsection{Experimental Results} 

\subsubsection{GF-RSSM supports accurate policy-conditioned prediction.}
\label{subsec:world_model_results}

Table~\ref{tab:world_model_results} evaluates deterministic prior rollouts on held-out synthetic episodes. GF-RSSM achieves the lowest error on all three admission metrics. Relative to the no-graph ablation, the full model reduces admission MAE by 26\% (0.211 to 0.156), RMSE by 29\% (0.311 to 0.221), and five-step cumulative error by 49\% (0.070 to 0.036). It similarly improves over Action-LSTM by 26\%, 29\%, and 36\%, respectively. Although uncertainty over three checkpoints limits strong statistical conclusions for MAE and RMSE, the cumulative-error improvement is consistent across checkpoints. 

\begin{figure}[t]
    \centering
    \includegraphics[width=\linewidth]{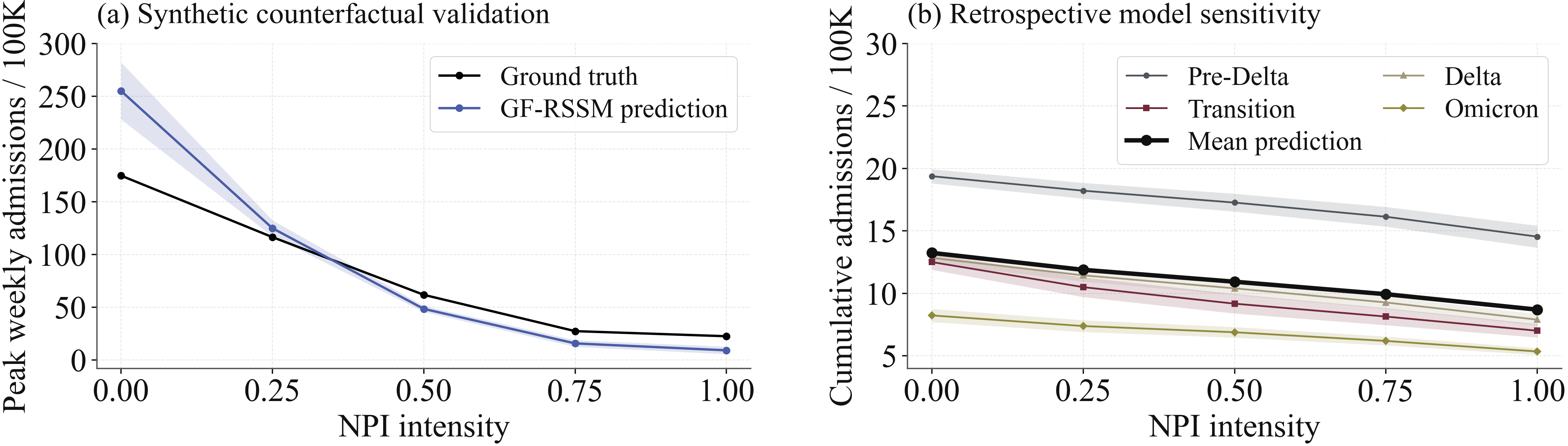}
    \caption{
    {Policy-conditioned admission response.} 
    {(a)} Peak weekly admissions under alternative NPI intensities applied from a common synthetic state. 
    {(b)} Predicted cumulative admissions under the same sweep at four held-out U.S.\ decision origins.}
    \label{fig:dose_response}
    \vspace{-5mm}
\end{figure}

Figure~\ref{fig:dose_response} examines whether this predictive accuracy extends to policy-conditioned responses. In simulation, GF-RSSM preserves the monotonic NPI dose ordering but overpredicts admissions at low NPI and underpredicts them at high NPI. On retrospective data, predicted admissions decrease with NPI across all four held-out decision origins. These real-data curves demonstrate stable model sensitivity, not causal effects, because counterfactual outcomes are unavailable.

\begin{table*}[t]
\centering
\caption{{World-model fidelity in the synthetic environment.} Values are mean \(\pm\) s.d. Admission errors are measured per 100K; cumulative error is over a five-step rollout. Lower is better.}
\label{tab:world_model_results}
\small
\setlength{\tabcolsep}{8pt}
\begin{tabular}{@{}lcccc@{}}
\toprule
\textbf{Model}
& \textbf{Adm.\ MAE} \(\downarrow\)
& \textbf{Adm.\ RMSE} \(\downarrow\)
& \textbf{Cum.\ err.\ @5} \(\downarrow\)
& \textbf{Params} \\
\midrule
\multicolumn{5}{@{}l}{\emph{Statistical baselines}} \\
Persistence
    & 1.563 \(\pm\) 0.053
    & 1.927 \(\pm\) 0.056
    & 0.089 \(\pm\) 0.003
    & 0 \\
Climatology
    & 1.129 \(\pm\) 0.021
    & 1.283 \(\pm\) 0.027
    & 0.528 \(\pm\) 0.011
    & 0 \\
Ridge \(+\) action
    & 0.341 \(\pm\) 0.008
    & 0.437 \(\pm\) 0.003
    & 0.068 \(\pm\) 0.006
    & 84 \\
VARX(1) \(+\) action
    & 0.329 \(\pm\) 0.008
    & 0.424 \(\pm\) 0.001
    & 0.066 \(\pm\) 0.007
    & 1,960 \\
\midrule
\multicolumn{5}{@{}l}{\emph{Learned dynamics models}} \\
Action-LSTM
    & 0.212 \(\pm\) 0.044
    & 0.310 \(\pm\) 0.062
    & 0.056 \(\pm\) 0.007
    & 25,507 \\
GF-RSSM (no graph)
    & 0.211 \(\pm\) 0.033
    & 0.311 \(\pm\) 0.046
    & 0.070 \(\pm\) 0.012
    & 92,972 \\
\midrule
\textbf{GF-RSSM (ours)}
    & \textbf{0.156 \(\pm\) 0.049}
    & \textbf{0.221 \(\pm\) 0.070}
    & \textbf{0.036 \(\pm\) 0.003}
    & 92,972 \\
\bottomrule

\vspace{-6mm}
\end{tabular}
\end{table*}

\subsubsection{Learned rollouts yields effective resources-feasible interventions.}
\label{subsec:planning_results} 

We evaluate all methods under the same health--intervention objective, action bounds, shared budgets, and final resource projection. Table~\ref{tab:planning_results} reports simulator-realized objective values across three intervention-cost regimes. The best constant-NPI policy saturates the shared budget, providing a strong non-adaptive comparator. EpiMind remains within 1.1\%, 1.5\%, and 4.9\% of this comparator at \(\lambda_{\mathrm{NPI}}\in\{3,10,30\}\), respectively, with paired regret \(0.93\pm1.10\) at \(\lambda_{\mathrm{NPI}}=10\). It also consistently outperforms PPO, MPC-SEIR, independent MPC, and the remaining planning baselines. All executed allocations have zero post-projection budget excess. These results show that planning through learned joint rollouts produces effective resource-feasible interventions. The learned-versus-oracle decomposition in Appendix~\ref{app:planning_extended} suggests a model contribution to the remaining regret, but the effect is not statistically resolved with four paired cells.

\begin{table*}[h]
\centering
\caption{
{Matched-objective planning performance.} Values are mean \(\pm\) s.d. Lower is better.}
\label{tab:planning_results}
\small
\setlength{\tabcolsep}{7pt}
\begin{tabular}{@{}lcccc@{}}
\toprule
& \multicolumn{3}{c}{\textbf{Objective \(J\)} \(\downarrow\)}
& \textbf{Regret at \(\lambda=10\)} \(\downarrow\) \\
\cmidrule(lr){2-4}
\textbf{Method}
& \(\lambda=3\)
& \(\lambda=10\)
& \(\lambda=30\)
& mean \(\pm\) s.d. \\
\midrule
Best feasible constant$^{\dagger}$
& \textbf{54.7} & \textbf{58.9} & \textbf{70.9}
& \textbf{0.00 \(\pm\) 0.00} \\
\midrule
PPO
& 60.4 & 64.6 & 76.6 & 5.69 \(\pm\) 3.71 \\
MPC-SEIR
& 108.6 & 112.8 & 124.6 & 53.9 \(\pm\) 30.9 \\
Independent MPC
& 366.3 & 431.1 & 394.1 & 372 \(\pm\) 307 \\
Greedy
& 7,186 & 7,189 & 7,197 & 7,130 \(\pm\) 1,221 \\
D-ADMM
& 9,300 & 9,263 & 9,224 & 9,204 \(\pm\) 583 \\
HRL
& 9,429 & 9,429 & 9,429 & 9,370 \(\pm\) 478 \\
No intervention
& 13,853 & 13,853 & 13,853 & 13,795 \(\pm\) 29 \\
\midrule
\textbf{\model{}}
& \textbf{55.3} & \textbf{59.8} & \textbf{74.4}
& \textbf{0.93 \(\pm\) 1.10} \\
\bottomrule

\end{tabular}

\vspace{2pt}
{\footnotesize
$^{\dagger}$Constant NPI at the shared-budget cap; no adaptive planning. \ \ \ \ \ \ \ \ \ \  \ \ \ \ \ \ \ \ \ \ \ \ \ \ \ \ \ \ \ }

    \vspace{-4mm}
\end{table*}

\subsubsection{Resource projection guarantees feasible execution.}
\label{subsec:constraint_results} 
\vspace{-1mm}

Table~\ref{tab:constraint_results} evaluates the projected actions that are executed. All allocations satisfy the encoded linear resource constraints, with zero maximum budget excess. Projection modifies synthetic proposals more than real-context proposals, as indicated by their mean displacement (\(0.476\) versus \(0.0039\)). After projection, all evaluated world-model rollouts satisfy the STL specification with positive exact robustness. 
Resource feasibility is guaranteed for the encoded linear constraints, whereas STL satisfaction is model relative and does not certify the unknown environment. Robustness to operational perturbations and epidemiological model mismatch is reported in Appendix~\ref{app:robustness} (Figure~\ref{fig:robustness}).

\vspace{-2mm}
\begin{table}[h]
\centering
\caption{{Constraint handling and model-relative verification.} }
\vspace{-1mm}
\label{tab:constraint_results}
\small
\setlength{\tabcolsep}{3.5pt}
\begin{tabular}{@{}lrrrrr@{}}
\toprule
\textbf{Setting}
& \textbf{Feasible} (\%)
& \textbf{Max excess}
& \textbf{Projection\ displacement}
& \textbf{STL satisfaction} (\%)
& \textbf{STL robustness} \\
\midrule
Synthetic & 100.0 & 0 & 0.476 \(\pm\) 0.201   & 100.00 & 0.0021 \(\pm\) 0.0004 \\
Real context  & 100.0 & 0 & 0.0039 \(\pm\) 0.0400 & 100.00 & 0.0016 \(\pm\) 0.0003 \\
\bottomrule
    \vspace{-4mm}
\end{tabular}
\end{table}

\subsubsection{Coordination benefits are modest under matched intervention burden.} 
\label{subsec:coordination_results}

Table~\ref{tab:realized_policy} evaluates realized outcomes under a common objective and shared resource constraints. Among the directly comparable planning methods, which incur nearly identical NPI burden (\(0.423\)--\(0.424\)), \model{} reduces \(J\) by 1.14\% relative to global-only ADMM, 2.13\% relative to graph-free planning, 2.21\% relative to time-shuffled planning, and 3.40\% relative to independent MPC. These results suggest benefits from graph structure and temporal allocation, although their magnitude is small. A stricter matched-burden ablation in Appendix Table~\ref{tab:coord_ablation} holds the NPI trajectory fixed step by step, isolating where interventions are allocated from how much is spent. Under this control, the coordination gains fall below 1\%, indicating that much of the uncontrolled difference arises from intervention burden rather than allocation alone. Region shuffling is reported only as a sensitivity diagnostic because shuffling after projection breaks the population-weighted budget constraints.
A matched-burden component ablation further isolates the coordination mechanism (Appendix Table~\ref{tab:coord_ablation}). The results indicate that the coordination signals provide consistent but incremental gains once intervention burden is controlled.

\begin{table*}[t]
\centering
\caption{
Realized policy performance in the real-context semi-synthetic evaluation. 
}
\vspace{-2mm}
\label{tab:realized_policy}
\small
\setlength{\tabcolsep}{7pt}
\begin{tabular}{@{}lrrrrr@{}}
\toprule
\textbf{Method}
& \textbf{Adm./100K} \(\downarrow\)
& \textbf{NPI burden}
& \(J^{\dagger}\) \(\downarrow\)
& \(\Delta J\) (\%)
& \textbf{\model{} wins (\(p\))} \\
\midrule
Standard ADMM
& 30.24 & 0.423 & 31.51 & \(+1.14\) & 23/36 (0.132) \\
Centralized MPC
& 30.65 & 0.424 & 31.92 & \(+2.45\) & 32/36 (\(<\!0.001\)) \\
Independent MPC
& 30.94 & 0.424 & 32.21 & \(+3.40\) & 26/36 (0.011) \\
Uniform allocation
& 31.23 & 0.479 & 32.67 & \(+4.87\) & 30/36 (\(<\!0.001\)) \\
EpidRLearn
& 35.64 & 0.500 & 37.14 & \(+19.23\) & 36/36 (\(<\!0.001\)) \\
PPO
& 35.69 & 0.499 & 37.19 & \(+19.37\) & 36/36 (\(<\!0.001\)) \\
Incidence-proportional
& 36.61 & 0.352 & 37.67 & \(+20.92\) & 30/36 (\(<\!0.001\)) \\
No intervention
& 64.14 & 0.000 & 64.14 & \(+105.90\) & 36/36 (\(<\!0.001\)) \\
\midrule
\textbf{\model{}}
& \textbf{29.89} & 0.423 & \textbf{31.15} & --- & --- \\
\bottomrule
\end{tabular}

\vspace{2pt}
{\footnotesize
\(^\dagger\)Policies are evaluated by \(J=\mathrm{Adm./100K}+\lambda_{\mathrm{NPI}}\mathrm{NPI}\); 
\(\Delta J\) is the percentage change relative to \model{}.\\
%
}
    \vspace{-4mm}
\end{table*}


\vspace{-1mm}
\subsubsection{\model{} transfers to real-context multi-region planning.} 
\label{subsec:real_data} 
\vspace{-1mm}

\begin{figure}[h]
\centering
\includegraphics[width=\linewidth]{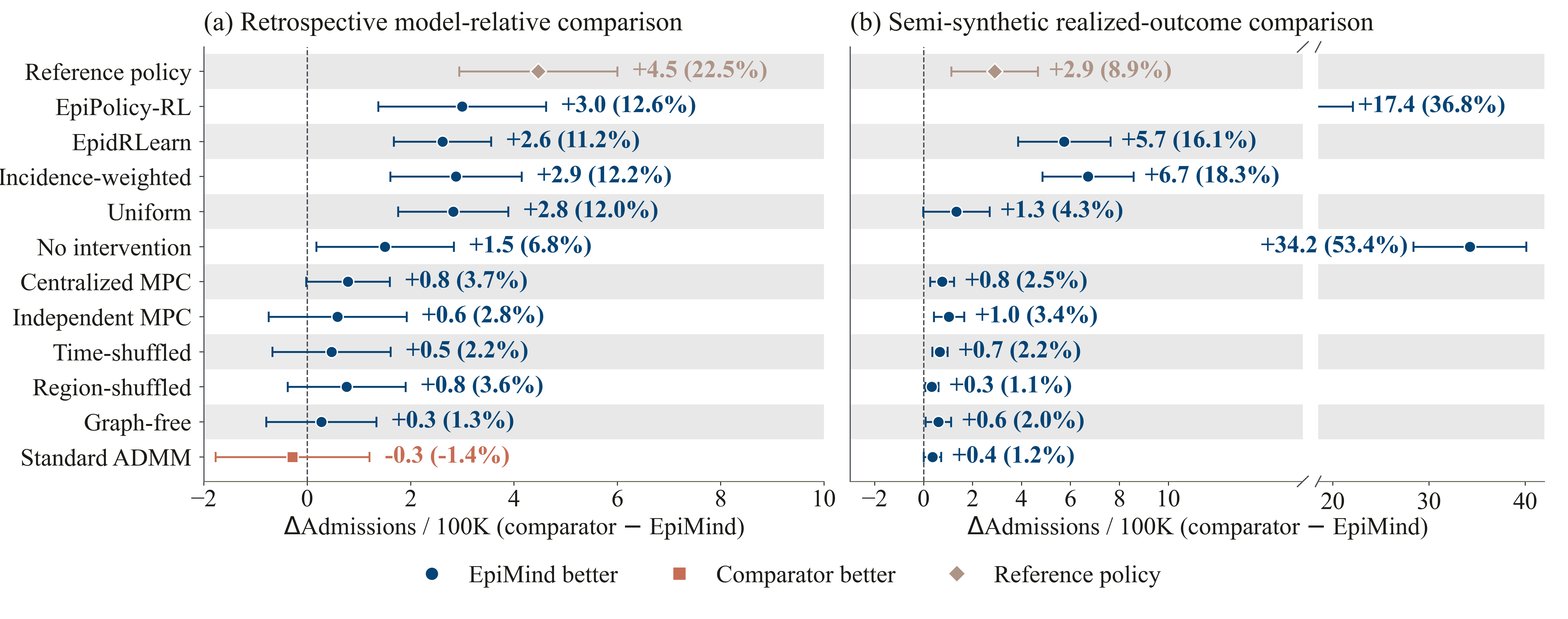}
\vspace{-8mm}
\caption{
{Real-context policy comparisons.} Each point reports the mean paired admission difference $\Delta\mathrm{Adm} =\mathrm{Adm}_{\mathrm{comparator}}-\mathrm{Adm}_{\model{}}$; positive values (blue circles) favor \model{}; horizontal bars denote 95\% confidence intervals. {(a)} Retrospective Track~A reports model-relative projections at held-out U.S.\ state-level decision origins. {(b)} Semi-synthetic Track~B reports realized outcomes under known simulation dynamics. Diamonds denote the track-specific reference policy. 
}
\label{fig:realcontext_comparison}
\vspace{-3mm}
\end{figure}

Figure~\ref{fig:realcontext_comparison} summarizes paired comparisons across both real-context tracks. In retrospective Track~A, \model{} projects fewer admissions than most comparators, although Standard ADMM is marginally better on average. These comparisons are model relative because the candidate policies were not executed. In semi-synthetic Track~B, where counterfactual outcomes are known, \model{} outperforms every deployable comparator. Its largest gains are over no intervention, EpiPolicy-RL, incidence-weighted allocation, and EpidRLearn. Smaller differences from Standard ADMM, graph-free planning, and shuffled controls indicate that constrained optimization provides most of the improvement, with graph and temporal coordination contributing incrementally.

\begin{figure}[t] 
\centering 
\includegraphics[width=\linewidth]{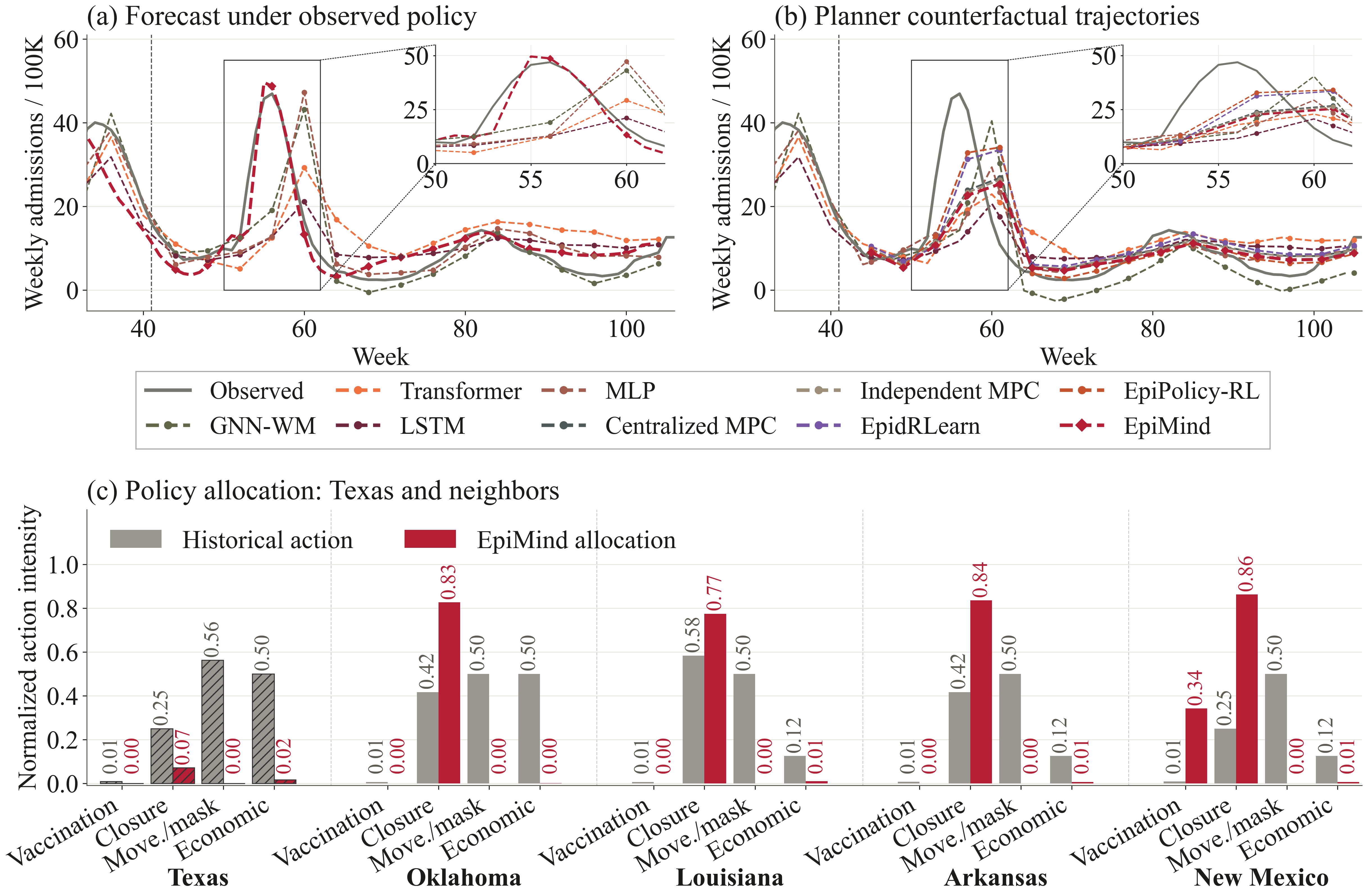} 
\vspace{-3mm}
\caption{ {Texas case study.} 
{(a)} Admission forecasts under the observed policy; the inset enlarges weeks 50--60. 
{(b)} Model-relative counterfactual trajectories under matched resource constraints; the observed trajectory provides context but is not an outcome of the unexecuted policies. 
{(c)} EpiMind's first-step allocation compared with historical actions in Texas and neighboring states. The dashed vertical line marks the training cutoff. MPC and RL baselines appear only in panel~(b). } 
\label{fig:texas_case} 
\vspace{-8mm}
\end{figure}

We next use Texas as a representative decision origin to illustrate the retrospective forecasting and planning workflow (Figure~\ref{fig:texas_case}, Figure~\ref{fig:texas_only}). Compared with the forecasting baselines, GF-RSSM more closely tracks the principal admission peak, while \model{} and the planning baselines produce distinct model-relative trajectories under matched constraints. 
\model{} assigns distinct actions to Texas and its mobility-connected neighbors under shared constraints (Figure~\ref{fig:texas_case}c)). The empirical mobility graph and additional state-level rollouts are shown in Appendix Figures~\ref{fig:mobility} and~\ref{fig:multistate_rollouts}.

\vspace{-1mm}
\section{Conclusion and Limitations} 
\label{sec:conclusion} 
\vspace{-2mm}

We presented \model{}, a graph world-model framework for constrained epidemic planning across regions. Its parameter-shared GF-RSSM maintains region-specific beliefs and predicts joint trajectories under candidate policies. GT-ADMM uses these rollouts to coordinate regional interventions and project shared allocations onto the feasible set. The experiments demonstrate accurate policy-conditioned prediction, effective planning through learned dynamics, and exact enforcement of the specified linear resource budgets. Under matched intervention burden, however, graph coordination provides a measurable but incremental benefit.

Several limitations remain. Projection guarantees only the constraints encoded in the feasible set, and STL satisfaction applies to learned trajectories rather than the unknown environment. Planning quality depends on world-model calibration, and the nonconvex GT-ADMM procedure has no global convergence guarantee. Because real-world outcomes under alternative policies are unobserved, the predicted trajectories and spillover effects represent model-based sensitivities rather than causally identified counterfactuals~\cite{hernan2020causal}. Finally, intervention costs and allocation priorities must reflect local economic, ethical, and public-health considerations. \model{} is therefore intended to support policy comparison and resource allocation, not to make decisions autonomously.

\vspace{-1mm}
\section{Software and Data}
\vspace{-2mm}
We release the full implementation at 
\url{https://anonymous.4open.science/r/epimind-9706/README.md}.

\newpage
\section*{AI Use Disclosure}

Generative AI tools were used to assist with literature retrieval and
discovery and to improve the clarity and readability of the manuscript.
All AI-assisted text was reviewed and revised by the authors, and all
citations and literature-derived statements were verified against their
original sources. The authors take full responsibility for the final
content of this work.

\bibliographystyle{iclr2027_conference}
\bibliography{epimind}

\newpage
\appendix
\setcounter{table}{0}
\renewcommand{\thetable}{A\arabic{table}}

\setcounter{figure}{0}
\renewcommand{\thefigure}{A\arabic{figure}}

\section{Epidemiological Foundations}
\label{sec:epi_foundations}

This section introduces the epidemiological structure underlying \model{}, including its latent dynamics, observations, and planning constraints.

\begin{figure}[h]
    \centering
     \includegraphics[width=.9\linewidth]{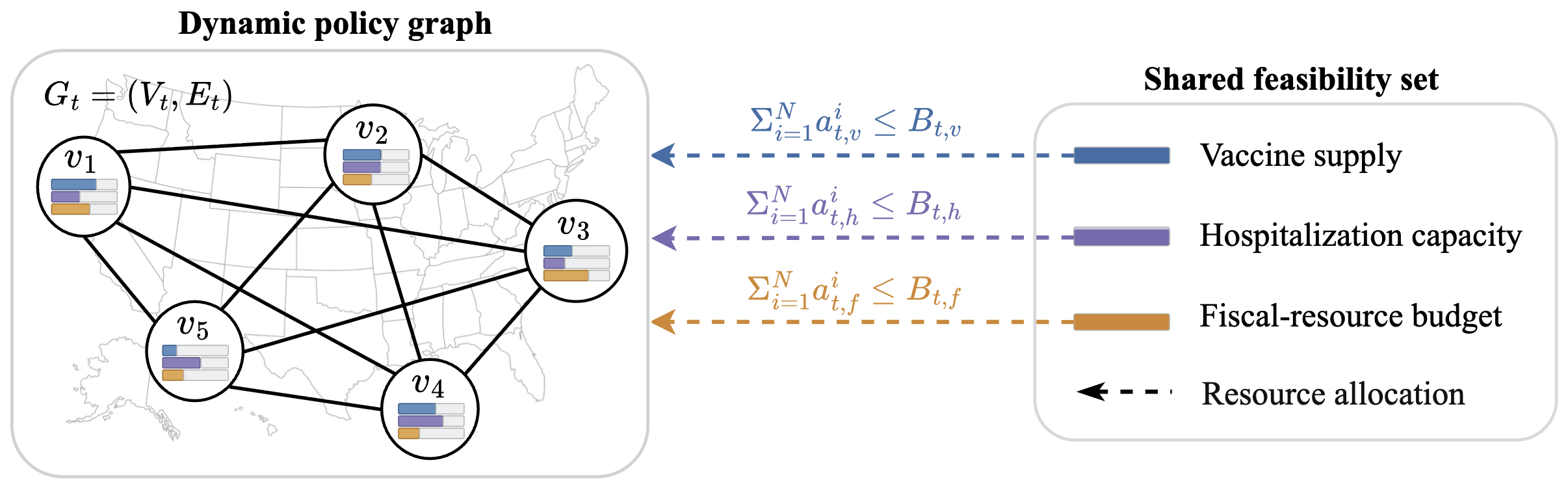}
        \caption{
        Multi-region policy graph under shared resource constraints.
        Regions are represented as nodes \(v_i\) in a dynamic policy graph \(G_t=(V_t,E_t)\), with edges encoding interregional mobility coupling.
        Each region receives allocations of vaccine, hospitalization-capacity, and fiscal resources. The joint regional allocations must satisfy graph-level resource budgets,
        \(\sum_{i=1}^{N} a_{t,k}^{i}\le B_{t,k}\), for
        \(k\in\{v,h,f\}\). 
        }
        \label{fig:epidemic_graph}
\end{figure}


\subsection{Epidemics as graphs.}
The spatiotemporal policy graph enables reasoning at three levels:

\begin{itemize}
\item \textbf{Node level.} Each node represents a region and encodes its latent state, healthcare capacity, population characteristics, and interventions through policy actions.
\item \textbf{Edge level.} Each edge encodes inter-regional coupling, including mobility flows, spatial proximity, or shared infrastructure that mediates epidemic spillovers and policy coordination between connected regions.
\item \textbf{Graph level.} Global resource constraints (vaccine supply, hospitalization capacity, fiscal-resource budget) and equity requirements operate over the entire graph, coupling all regions' feasible action sets.
\end{itemize}

\paragraph{Latent states. }
As the true compartmental state is \emph{never directly observed}, surveillance systems report cases (a function of testing), hospitalizations (a function of severity and care-seeking), and deaths (a lagged indicator). Each of these is a noisy, delayed, and policy-dependent projection of the underlying state. For instance, expanding testing increases reported cases without changing true cases, while reducing testing has the opposite effect. This means that an observed decline in cases may reflect either genuine transmission reduction or reduced testing coverage, a fundamental ambiguity that forecasting models cannot resolve without separating latent state from observation.
This motivates the use of a \emph{latent state representation} that captures the true epidemic reality, including infection burden, immunity, behavioral compliance, and variant fitness behind the noisy observations~\cite{memon2026toward}. By separating latent dynamics from a policy-dependent observation model, this representation disentangles genuine transmission changes from surveillance artifacts, enabling policy reasoning based on inferred reality rather than distorted measurements.

\paragraph{Interventions, behavioral mediation, and delayed effects. }
Epidemic interventions include NPIs, PIs, and surveillance interventions. A critical feature of these interventions is \emph{behavioral mediation}~\cite{funk2010modelling}. Mandates do not directly reduce transmission. Instead, they alter human behavior, which in turn changes contact patterns and infection risk. Compliance with interventions is partial, heterogeneous, and time-varying, depending on perceived risk, fatigue, trust, and economic pressure.
In addition, all interventions operate with \emph{temporal delays}. An NPI enacted today cannot reduce transmission until behavioral changes propagate through the population (typically 1--2 weeks, corresponding to the generation interval $\tau_g$). Vaccination requires weeks to build immunity. These delays mean that the effect of an action taken at time $t$ does not appear in surveillance data until $t + \tau_g$ or later, so a planner cannot validate or course-correct an active intervention against current observations. By the time an intervention's effect becomes visible, the epidemic has already evolved into a different state, which is precisely why planning under epidemic dynamics must rely on forward simulation of latent state under candidate interventions rather than on real-time feedback from surveillance alone.

\paragraph{Resource constraints. }
Since the resources available for intervention are \emph{shared across regions and physically finite}, epidemic policy can be formulated as a constrained optimization, with two properties that distinguish it from standard constrained programs. First, feasibility must hold at every decision step, since vaccine doses cannot be administered beyond the available stock at time $t$, hospital and ICU occupancy cannot exceed bed capacity, and per-period public-health expenditures cannot exceed the appropriated budget. Formulations that incorporate constraints as additive cost terms drive expected violation to zero only asymptotically. They are insufficient here, because an over-allocation at any single step cannot be implemented in practice. Second, each constraint is a sum across regions rather than a per-region limit, as the underlying resources are pooled at the national or system-wide level. Therefore, each region's feasible action set depends on what the others request, and the optimization cannot be decomposed into independent single-region problems.

\subsection{From Policy-Conditioned Prediction to Constrained Coordination}
\label{subsec:planning_pipeline}

The epidemiological properties described in Section~\ref{sec:epi_foundations} create the following unique computational challenges that current methods only partially address:

    

    

\begin{enumerate}[label=(\Alph*)]
\item \textbf{Policy-conditioned prediction under uncertainty.} Epidemic dynamics are nonlinear, partially observed, and shaped by latent behavioral responses~\cite{su2026how}. Planning therefore requires inferring regional latent states from noisy surveillance and generating joint trajectories under candidate interventions. 

\item \textbf{Coordination under shared constraints.} Regional decisions are coupled by mobility-driven spillovers, finite shared resources, and delayed intervention effects~\cite{flaxman2020estimating}. Effective planning must account for these interactions while enforcing resource limits and temporal requirements explicitly. 

\item \textbf{Verification after projection.} Resource projection can alter the planner's unconstrained proposal. Specifications must therefore be evaluated on a new joint rollout under the projected action that will be executed. This separates exact resource feasibility from model-relative temporal verification.

\end{enumerate}

\begin{table}[h]
\small
\centering
\caption{Epidemiological challenges and their computational treatment in \model{}.}
\label{tab:why_epimind}
\small
\begin{tabular}{p{1.1cm}p{3.8cm}p{2.2cm}p{5cm}}
\toprule
\textbf{Challenge} & \textbf{Epidemiological property} & \textbf{Component} & \textbf{Computational response} \\
\midrule
\multirow{4}{*}{\parbox{3.2cm}{(A)}}
 & Latent disease burden & GF-RSSM & Latent belief inference \\
 & Policy-dependent surveillance & GF-RSSM & Action-conditioned observation model \\
 & Delayed intervention effects & GF-RSSM & Multi-horizon policy-conditioned rollout \\
 & Cross-region spillovers & GF-RSSM & Graph attention over neighbors \\
\midrule
\multirow{4}{*}{\parbox{3.2cm}{(B)}}
 & Shared resource limits & GT-ADMM & Capped-simplex projection \\
 & Cross-border coordination & GT-ADMM & Neighbor-consensus signal \\
 & Spillover externalities & GT-ADMM & Spillover-sensitivity accumulator \\
 & Temporal specifications & GT-ADMM/STL & Differentiable robustness objective \\
\midrule
\parbox{3.2cm}{(C)}
 & Proposal--execution mismatch & Verification & Joint reroll under projected action \\
\bottomrule
\end{tabular}
\end{table}

These challenges motivate the pipeline in Figure~\ref{fig:pipeline}. The learned graph world model maps the current posterior beliefs and a candidate joint action sequence to a differentiable joint trajectory. GT-ADMM uses these trajectories as its predictive objective, updates region-specific action sequences, and projects the joint allocation onto the shared-resource polytope. Finally, the framework rerolls the projected action and evaluates exact STL robustness under the learned model. Dynamics learning remains predictive: constraint satisfaction is not inserted into the world-model training loss. The interaction between learning and planning occurs operationally, because executed actions determine the observations used for the next posterior update.

Table~\ref{tab:why_epimind} summarizes the division of computational responsibilities. The GF-RSSM answers how the coupled regional epidemic is predicted to evolve under a candidate joint intervention. GT-ADMM answers which region-specific intervention sequence minimizes the predicted objective while respecting the specified shared budgets. Post-projection reroll distinguishes the proposed trajectory from the model-predicted trajectory under the action that will actually be executed. This modular separation permits the dynamics model to be trained for predictive fidelity and the planner to enforce operational constraints explicitly, without claiming that optimization constraints reshape the learned dynamics.

\section{\model{} Framework}
\label{sec:framework}

\begin{algorithm}[h]
\caption{\model{} constrained receding-horizon planning}
\label{alg:epimind}
\begin{algorithmic}[1]
\Require Frozen world model \(\mathcal M_\theta\), graph \(G_t\), observation
\(\mathbf o_t\), previous plan \(\mathbf a^\star_{t-1}\), feasible set
\(\mathcal Z_t\), temporal specifications \(\Phi\), horizon \(H\)
\Ensure Executed joint action \(\mathbf a_t^\star\)

\State Update regional beliefs:
\(
(\mathbf b_t,\mathbf z_t)
\gets
\operatorname{Posterior}_{\theta}
(\mathbf o_t,\mathbf a_{t-1},G_t)
\)

\State Warm-start action sequences
\(
\mathbf a^0
\gets
\operatorname{Shift}(\mathbf a^\star_{t-1})
\);
set
\(
\mathbf y^0\gets\mathbf a^0
\)
and
\(
\boldsymbol\gamma^0,\mathbf u^0,\boldsymbol\eta^0\gets 0
\)

\For{\(k=0,\ldots,K_{\max}-1\)}
    \ForAll{regions \(i\) \textbf{in parallel}}
        \State Update \(a^{i,k+1}\) by minimizing~\eqref{eq:x_step}
        through a joint prior rollout
        \[
        \hat{\mathbf x}_{t+1:t+H}
        =
        \mathcal M_\theta
        \!\left(
        \mathbf b_t,\mathbf z_t,
        a^i,\mathbf a^{-i,k};G_t
        \right)
        \]
    \EndFor

    \State Project shared allocations:
    \[
    \mathbf y^{k+1}
    \gets
    \operatorname{Proj}_{\mathcal Z_t}
    \!\left(\mathbf a^{k+1}+\mathbf u^k\right)
    \]

    \State Update
    \(
    \boldsymbol\gamma^{k+1},
    \mathbf u^{k+1},
    \boldsymbol\eta^{k+1}
    \)
    using~\eqref{eq:edge_dual}--\eqref{eq:spillover_update}

    \If{primal residuals satisfy their tolerances}
        \State \textbf{break}
    \EndIf
\EndFor

\State Reroll the proposal \(\mathbf a^K\) and projected allocation
\(\mathbf y^K\) through the same joint world model

\State Evaluate exact model-relative robustness:
\[
\rho_{\mathrm{exec}}
=
\rho_{\mathrm{exact}}
\!\left(
\Phi,\mathcal M_\theta(\mathbf y^K;G_t)
\right)
\]

\State Execute the first projected action:
\(
\mathbf a_t^\star=\mathbf y_t^K
\)

\State \Return
\(
\mathbf a_t^\star,\rho_{\mathrm{exec}}
\)
\end{algorithmic}
\end{algorithm}

\subsection{Signal Temporal Logic for Epidemiological Rules}
\label{app:stl}

We use differentiable STL robustness during optimization and exact robustness for model-relative verification of the projected rollout. In the experiments, the specification requires predicted hospital occupancy to remain below regional capacity over the planning horizon:

\begin{equation}
\Phi^i
=
\Box_{[0,H]}
\left(
\widehat H_{t+h}^i
\leq
H_{\mathrm{cap}}^i
\right).
\label{eq:stl_capacity}
\end{equation}

A positive exact robustness score certifies satisfaction only on the learned rollout; it is not a guarantee for the unknown environment.

\section{Experimental settings}
\label{sec:setting}

\subsection{Real-World Data Sources and Preprocessing} 
\label{app:real_data} 

We construct a weekly U.S.\ state-level panel from the following sources: 
\textit{(1) Epidemic burden and hospital resources} are obtained from the HHS \href{https://healthdata.gov/dataset/COVID-19-Reported-Patient-Impact-and-Hospital-Capa/6xf2-c3ie} {COVID-19 Reported Patient Impact and Hospital Capacity by State Timeseries}, which provides new COVID-19 admissions, inpatient and ICU occupancy, staffed beds, and ICU capacity. 
\textit{(2) Reported infections and deaths} are obtained from the CDC \href{https://data.cdc.gov/}{COVID-19 state surveillance datasets}; these measure reported cases rather than latent infections and are treated as noisy observations. 
\textit{(3) Vaccination} is drawn from the CDC \href{https://data.cdc.gov/d/unsk-b7fc} {COVID-19 Vaccinations in the United States, Jurisdiction} dataset, including doses delivered and administered, primary-series completion, and booster coverage. 
\textit{(4) Policy interventions} are taken from the \href{https://statepolicies.com/}{COVID-19 U.S.\ State Policy Database (CUSP)} (\href{https://github.com/USCOVIDpolicy/COVID-19-US-State-Policy-Database} {GitHub}), which records state-level mask requirements, gathering limits, stay-at-home orders, school and business closures, emergency declarations, and economic-support policies. We aggregate active mitigation policies into a normalized NPI-intensity index and use the economic-policy fields as fiscal-support indicators. 
\textit{(5) Interregional mobility} is obtained from \href{https://advanresearch.com/products/patternsplus}{Advan Patterns+} (\href{https://docs.advanresearch.com/}{data dictionary}), whose visitor origins are aggregated into directed state-to-state flows defining the time-varying graph \(G_t\). 
\textit{(6) State populations} are obtained from the U.S.\ Census Bureau's \href{https://data.census.gov/table/PEPPOP2021.NST_EST2021_POP} {2021 Population Estimates Program} and are used to calculate per-100K outcomes and population-weighted allocations.

\subsection{Baseline Methods} 
\label{app:baselines} 
We organize the baselines by the capability being evaluated. 

\paragraph{World-model baselines.} 
\emph{Persistence} repeats the latest observation, whereas 
\emph{climatology} predicts the training-set mean. 
\emph{Ridge+action} and \emph{VARX(1)+action} are deterministic linear predictors conditioned on regional interventions. 
\emph{Action-LSTM} is a recurrent sequence model conditioned on the joint action vector~\cite{hochreiter1997long}. 
\emph{GF-RSSM (no graph)} retains the recurrent latent-state architecture but removes graph attention, isolating the contribution of mobility-based message passing. All forecasting models use the same train/test split and are evaluated in the same observation units and rollout protocol. 

\paragraph{Planning baselines.} 
\emph{No intervention} applies zero intervention throughout the horizon. 
\emph{Best feasible constant} selects a time-invariant NPI level by grid search under the same budget and evaluation objective. 
\emph{Uniform} divides each shared resource equally across regions, while \emph{incidence-weighted} allocates resources in proportion to current reported incidence. 
\emph{Greedy} assigns resources according to the immediate predicted health benefit without multi-step optimization. 
\emph{Centralized MPC} jointly optimizes all regional actions through a common finite-horizon objective, whereas \emph{independent MPC} optimizes each region without cross-region coordination. 
\emph{MPC-SEIR} plans with access to the simulator's compartmental state and fixed epidemiological parameters and is therefore an informed simulator-based comparator rather than a deployable real-data method. 
\emph{Standard ADMM} retains only the global resource-consensus variable and projection of conventional distributed ADMM~\cite{boyd2011admm}. 
\emph{D-ADMM} provides a distributed optimization baseline without EpiMind's learned graph coordination. 
The \emph{graph-free} ablation removes mobility coupling from EpiMind, while \emph{independent planning} removes both graph coupling and cross-region coordination. Region- and time-shuffled controls preserve intervention burden while disrupting spatial or temporal allocation. 

\paragraph{Policy-learning baselines.} We compare with PPO~\cite{schulman2017proximal}, MAPPO~\cite{yu2022mappo}, hierarchical reinforcement learning (HRL), and adapted implementations of EpidRLearn and EpiPolicy-RL. These methods are trained under the same action bounds and resource budgets; their outputs are passed through the same final projection used for EpiMind. Because the adapted epidemic-policy baselines do not use the authors' original implementations, we treat them as representative algorithmic adaptations rather than exact reproductions of published results. 

\paragraph{Oracle reference.} 
\emph{Oracle dynamics} uses the same planning interface as EpiMind but replaces the learned rollout with the hidden simulator dynamics. It measures the effect of dynamics-model error and is reported as an idealized reference, not as a deployable baseline.

\subsection{Parameter Table}
\label{tab:simulator_parameters}

Table~\ref{tab:params} summarizes the epidemiological, intervention, observation, world-model, planner, budget, objective, training, and evaluation parameters used in the experiments, together with their values and supporting references or implementation rationale.

\begin{table}[h]
\centering
\caption{Parameter inventory.}
\label{tab:params}
\small
\begin{tabular}{@{}p{2cm}p{3.8cm}p{1.8cm}p{5cm}@{}}
\toprule
\textbf{Symbol} & \textbf{Description} & \textbf{Value}
& \textbf{Reference / Justification} \\
\midrule
\multicolumn{4}{@{}l@{}}{\textit{Epidemiological}}  \\
$\beta$ & Transmission rate & 0.25/day & \citet{park2020reproduction}; $R_0=\beta/\gamma=2.5$ \\
$\sigma$ & E$\to$I rate & 0.25/day (4\,d) & \citet{li2020early}; \citet{lauer2020incubation} \\
$\gamma$ & I$\to$R rate & 0.1/day (10\,d) & \citet{he2020temporal} \\
$\delta_H$ & Hospitalization rate & 0.15 & \citet{garg2020hospitalization} (pre-Omicron) \\
$\delta_D$ & Base death rate & 0.01 & \citet{meyerowitzkatz2020systematic} \\
\midrule
\multicolumn{4}{@{}l@{}}{\textit{Intervention effects}} \\
NPI $\beta$-red. & Max local NPI effect & 60\% & \citet{brauner2021inferring}: 13--77\% \\
NPI import red. & Max effect on imported force & 40\% & Travel restriction is partial \\
FOI split & Local\,:\,imported weight & 0.7\,:\,0.3 & {Mobility coupling strength}\\
Vacc.\ uptake & $S\to R$ rate at full allocation & 0.01/day of $S$ & Vaccinated S move directly to $R$ \\
Fiscal scale & Funding$\to$compliance & 0.3 & \citet{hale2021global} \\
\midrule
\multicolumn{4}{@{}l@{}}{\textit{Capacity and overflow}} \\
$\text{LOS}$ & Hospital length of stay & 7 days & Converts incidence to occupancy \\
$m_{\max}$ & Max overflow mortality mult. & 5.0 & $\delta_D^{\text{eff}}=\min(\delta_D\,m,1)$ \\
$c_{\text{base}}, c_{\text{surge}}$ & Capacity fractions of pop. & 0.001, 0.002 & Baseline plus surge at full allocation \\
\midrule
\multicolumn{4}{@{}l@{}}{\textit{Observation model}} \\
$\sigma_{\text{obs}}$ & Measurement noise std & 0.005 & Additive Gaussian on all channels \\
$p_{\text{detect}}$ & Base detection prob. & 0.7 & \citet{wu2020estimating}: 14--86\% \\
NPI det.\ boost & Testing scales with NPI & $+0.2$, $[0.3,1]$ & \citet{bilinski2021passing} \\
$\tau_{\text{delay}}$ & Reporting delay & 1 week & Infection-to-confirmation lag \\
\midrule
\multicolumn{4}{@{}l@{}}{\textit{Graph world model (GF-RSSM)}} \\
$d_b, d_z, d_c$ & Belief / state / context & 64 / 16 / 32 & \citet{hafner2025mastering}; 5-region problem \\
$n_{\text{heads}}$ & Graph attention heads & 4 & \\
$d_o$ & Observation channels & 7 & \\
$\sigma_{\min}$ & Min prior/posterior std & 0.01 & \\ 
\midrule
\multicolumn{4}{@{}l@{}}{\textit{GT-ADMM coordinator}} \\
$\rho_e, \rho_g$ & Edge / global penalties & 1.0 & \citet{boyd2011distributed} (ADMM default) \\
$\sigma_{\text{prox}}$ & Proximal weight & 0.1 & \citet{wang2019randomized} \\
$K_{\max}$ & Max ADMM iterations & 15 & \\
inner steps & Gradient steps per $x$-step & 3 & \\
$\eta_x$ & $x$-step learning rate & 0.02 & \\
$\beta_{\text{STL}}$ & Smooth-STL temperature & 0.5 & Smooth for gradients \\
\midrule
\multicolumn{4}{@{}l@{}}{\textit{Shared budgets}} \\
$B_{\text{vac}}, B_{\text{hosp}}, B_{\text{fisc}}$ & Resource budgets & 3.0, 2.5, 3.5 & 0.6 / 0.5 / 0.7 per region \\
$B_{\text{NPI}}$ & Shared NPI budget & 3.0 & 0.6 per region \\
\midrule
\multicolumn{4}{l}{\textit{Objective}} \\
$\lambda_{\text{NPI}}$ & NPI cost weight & 0.1/3/\{3,10,30\} & Track A / Track B / synthetic sweep \\
$\lambda_{\text{res}}$ & Resource cost weight & 0.02 & Prices the budget channels \\
\midrule
\multicolumn{4}{@{}l@{}}{\textit{Training and evaluation}} \\
$H_{\text{rollout}}$ & Planning rollout horizon & 4 steps & CDC 4--6 week \citep{howerton2023evaluation} \\
$T$ & Episode length & 26 weeks & \\
Episodes & Train / val / test & 128 / 32 / 32 & Chronological split \\
Epochs & Max, with early stopping & 1000 & Validation-based selection \\
Seeds & Training / evaluation & 3 / 5 & \\
\bottomrule
\end{tabular}
\end{table}

\section{Results}

\subsection{World Model Evaluation}
\label{app:wm_extended}

\begin{figure}[h]
    \centering
    \includegraphics[width=\linewidth]
    {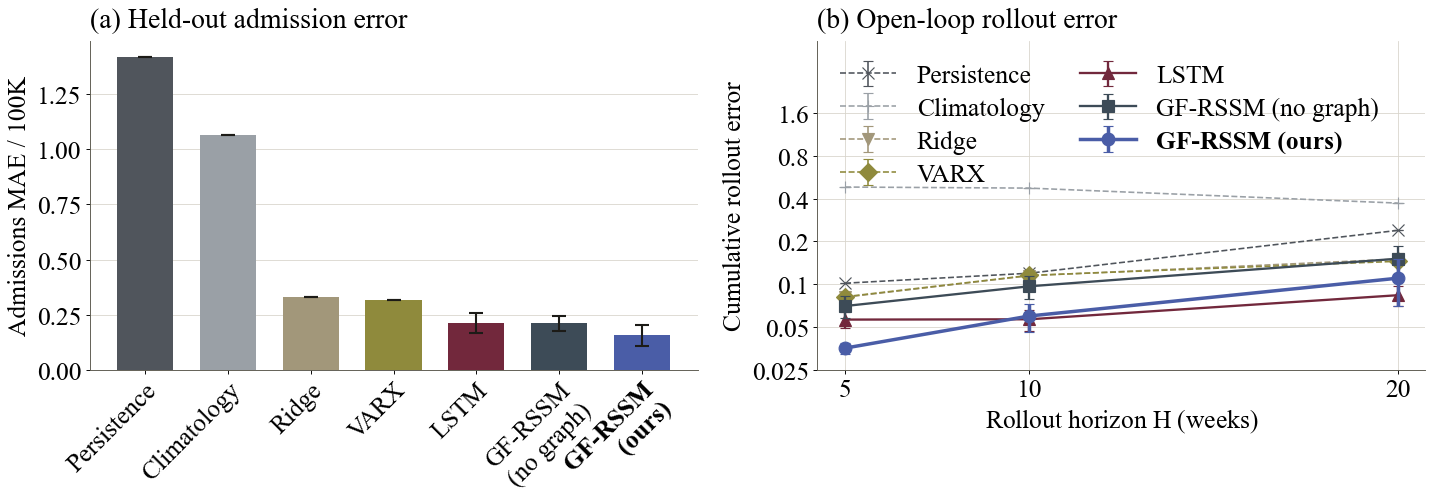}
    \caption{
    {Extended world-model evaluation on the synthetic benchmark.} 
    {(a)} Held-out admission MAE for learned architectures and statistical baselines under the same evaluation protocol. Error bars for learned models show mean \(\pm\) s.d.\ over three training seeds. 
    {(b)} Relative cumulative admission error over open-loop rollout horizons \(H\in\{5,10,20\}\), shown on a logarithmic scale. }
    \label{fig:wm_appendix}
\end{figure}

Figure~\ref{fig:wm_appendix} separates one-step admission accuracy from cumulative open-loop error. The full GF-RSSM achieves the lowest admission MAE and the lowest cumulative error at \(H=5\). Its advantage narrows by \(H=10\), and the action-conditioned LSTM performs better at \(H=20\), indicating greater long-horizon drift in the GF-RSSM. Removing graph attention consistently increases rollout error, while all learned models outperform the persistence and climatology references.

\subsection{Robustness Analysis}
\label{app:robustness}

\begin{figure}[h]
    \centering
\includegraphics[width=\linewidth]{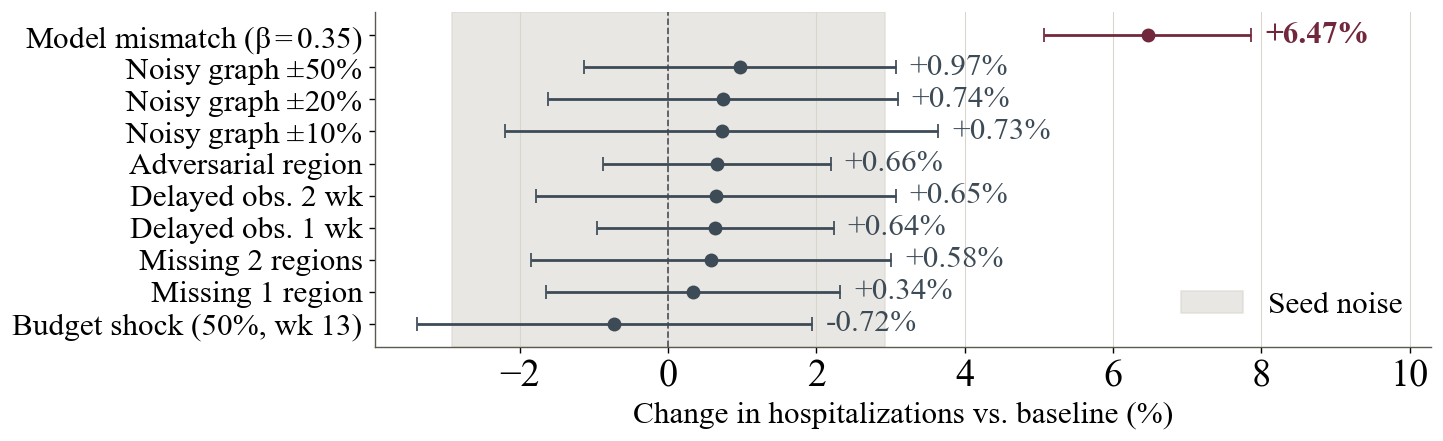}
    \caption{{Sensitivity to operational and model perturbations.} Changes in cumulative hospitalizations relative to each seed-matched baseline, reported as mean \(\pm\) s.d. The shaded band shows the largest within-condition standard deviation among the operational perturbations. Graph noise, masked regional observations, reporting delays, a mid-horizon budget cut, and regional non-compliance remain within this descriptive variability band. Epidemiological model mismatch produces the only substantially larger mean degradation.}

    \label{fig:robustness}
\end{figure}

Figure~\ref{fig:robustness} evaluates sensitivity to graph noise, missing or delayed observations, a mid-horizon budget cut, regional non-compliance, and epidemiological model mismatch. Across five paired seeds, the operational perturbations change cumulative hospitalizations by less than 1\% on average and remain within seed-level variability. Model mismatch produces the largest degradation (\(+6.47\%\)), identifying misspecified epidemic dynamics as the dominant tested failure mode. Given the limited number of seeds, these results are descriptive rather than evidence of statistical equivalence.

\subsection{Planning and Calibration Analysis}
\label{app:planning_extended}

Table~\ref{tab:oracle_decomp_full} expands the RQ2 regret analysis across intervention costs \(\lambda\) and planning horizons \(H\). Learned- and oracle-dynamics planners use the same optimizer and per-cell best-constant comparator. The model effect is therefore the paired regret difference \(R_{\mathrm{learned}}-R_{\mathrm{oracle}}\), isolating the change produced by replacing the learned rollout with the simulator dynamics.

\begin{table}[h]
\centering
\caption{{Learned- versus oracle-dynamics planning regret.} Values are mean \(\pm\) s.d.}
\label{tab:oracle_decomp_full}
\small
\setlength{\tabcolsep}{4pt}
\begin{tabular}{@{}ccrrrc@{}}
\toprule
\(\lambda\) & \(H\)
& \multicolumn{2}{c}{\textbf{Regret} \(\downarrow\)}
& \textbf{Model effect} \(\downarrow\)
& \(p^\dagger\) \\
\cmidrule(lr){3-4}
& & Learned & Oracle & & \\
\midrule
3  & 4  & 32.5 \(\pm\) 32.8   & 27.8 \(\pm\) 3.2
         & 4.7 \(\pm\) 34.6    & 1.000 \\
3  & 8  & 38.2 \(\pm\) 50.0   & 17.6 \(\pm\) 0.7
         & 20.5 \(\pm\) 50.3   & 1.000 \\
3  & 12 & 32.5 \(\pm\) 34.2   & 20.6 \(\pm\) 0.2
         & 11.9 \(\pm\) 34.2   & 1.000 \\
\midrule
10 & 4  & 217.0 \(\pm\) 229.3 & 103.2 \(\pm\) 3.8
         & 113.9 \(\pm\) 231.3 & 1.000 \\
10 & 8  & 184.9 \(\pm\) 201.1 & 79.3 \(\pm\) 0.4
         & 105.6 \(\pm\) 200.9 & 1.000 \\
10 & 12 & 186.4 \(\pm\) 198.0 & 78.2 \(\pm\) 3.1
         & 108.2 \(\pm\) 199.7 & 0.625 \\
\midrule
30 & 4  & 593.9 \(\pm\) 517.6 & 328.6 \(\pm\) 6.8
         & 265.4 \(\pm\) 513.3 & 0.625 \\
30 & 8  & 457.9 \(\pm\) 401.8 & 232.7 \(\pm\) 15.0
         & 225.2 \(\pm\) 392.6 & 0.625 \\
30 & 12 & 459.1 \(\pm\) 440.5 & 235.9 \(\pm\) 27.9
         & 223.2 \(\pm\) 424.9 & 0.625 \\
\bottomrule
\end{tabular}

\vspace{2pt}
{\footnotesize
\(\dagger\) \(p\) is an exact two-sided sign test. \ \ \ \ \ \ \ \ \ \ \ \ \ \ \ \ \ \ \ \ \ \ \ \ \ \ \ \ \ \ \ \ \ \ \ \ \ \ \ \ \ \ \ \ \ \ \ \ }
\end{table}

The learned planner has higher mean regret in every configuration, but its variation across cells is large: the model-effect standard deviation exceeds its mean in every row, and no sign test is significant. The grid therefore suggests a rollout-model contribution to regret but does not establish its magnitude at this sample size.

Table~\ref{tab:calibration_control_full} tests whether the dedicated hospitalization head and its calibration terms improve planning while holding the planner and comparator fixed. Because all variants are scored against the same per-cell best-constant policy, comparisons are paired.

\begin{table}[h]
\centering
\caption{Calibration ablation. Values are mean \(\pm\) s.d.\ over \(n{=}9\) cells.}
\label{tab:calibration_control_full}
\small
\setlength{\tabcolsep}{4pt}
\begin{tabular}{@{}c rrr cc cc@{}}
\toprule
& \multicolumn{3}{c}{\textbf{Regret} \(\downarrow\)}
& \multicolumn{2}{c}{\textbf{vs.\ decoder}}
& \multicolumn{2}{c}{\textbf{vs.\ uncalibrated}} \\
\cmidrule(lr){2-4}\cmidrule(lr){5-6}\cmidrule(lr){7-8}
\(\lambda\) & Calibrated & Shared dec. & Uncalib.
& cells & \(p\)
& cells & \(p\) \\
\midrule
3  & 56.5 \(\pm\) 50.5  & 72.8 \(\pm\) 85.4   & 68.8 \(\pm\) 42.2  & 5/9 & 1.000 & 7/9 & 0.180 \\
10 & 252.8 \(\pm\) 242.3 & 705.3 \(\pm\) 586.5 & 364.9 \(\pm\) 309.3 & 8/9 & \textit{0.039} & 9/9 & \textit{0.004} \\
30 & 641.4 \(\pm\) 474.1 & 2401.9 \(\pm\) 906.1 & 882.1 \(\pm\) 634.7 & 9/9 & \textit{0.004} & 7/9 & 0.180 \\
\bottomrule
\end{tabular}
\end{table}

At \(\lambda{=}10\), the calibrated head reduces mean regret by \(64.2\%\) relative to the shared decoder and by \(30.7\%\) relative to the uncalibrated head, with both paired tests significant. At \(\lambda{=}30\), it reduces mean regret by \(73.3\%\) relative to the shared decoder, but the additional benefit over the uncalibrated head is not significant. Neither contrast is established at \(\lambda{=}3\). These results support the dedicated head and calibration terms at intermediate intervention costs, while showing that aggregate mean ratios should not be interpreted as uniform per-cell gains.

\subsection{Coordination Ablation}
\label{app:coord_ablation}

We isolate the contributions of neighbor consensus \(\gamma\), predicted spillover sensitivity \(\eta\), and global resource consensus \(\mu\) under a matched-burden protocol. Every variant follows the full model's stepwise NPI-burden trajectory, so differences reflect where and when interventions are allocated rather than total NPI use.

\begin{table}[h]
\centering
\caption{{Matched-burden coordination ablation.}
Simulator-realized outcomes on the synthetic benchmark. All variants follow the reference configuration's stepwise NPI-burden trajectory. Here, \(\gamma\) denotes neighbor consensus, \(\eta\) spillover sensitivity, and \(\mu\) global resource consensus. \(\Delta\) is the percentage change relative to the full model; lower is better.}
\label{tab:coord_ablation}
\small
\setlength{\tabcolsep}{5pt}
\begin{tabular}{@{}lccc rr@{}}
\toprule
\textbf{Variant}
& \(\boldsymbol{\gamma}\)
& \(\boldsymbol{\eta}\)
& \(\boldsymbol{\mu}\)
& \textbf{Cum.\ adm./100K} \(\downarrow\)
& \(\boldsymbol{\Delta}\) (\%) \\
\midrule
Graph-free
& -- & -- & --
& 662.57 & \(+0.88\) \\

No edge (\(\gamma\) off)
& -- & \checkmark & \checkmark
& 659.87 & \(+0.47\) \\

No spillover (\(\eta\) off)
& \checkmark & -- & \checkmark
& 660.35 & \(+0.54\) \\

No global (\(\mu\) off)
& \checkmark & \checkmark & --
& 662.99 & \(+0.94\) \\
\midrule
\textbf{\model{}}
& \checkmark & \checkmark & \checkmark
& \textbf{656.81} & --- \\
\bottomrule
\end{tabular}
\end{table}

Full \model{} achieves the lowest cumulative admissions (656.81/100K), but the matched-burden gains are limited. Removing global resource consensus (\(\mu\)) produces the largest individual degradation (\(+0.94\%\)), followed by removing spillover sensitivity (\(\eta\), \(+0.54\%\)) and neighbor consensus (\(\gamma\), \(+0.47\%\)). Disabling all three signals increases admissions by \(0.88\%\). Thus, the coordination components provide complementary but incremental improvements once intervention burden is controlled; the substantially larger differences observed without burden matching partly reflect variation in total intervention effort rather than coordination alone.

\subsection{Real Mobility Graph}
\label{app:mobility}

\begin{figure}[h]
    \centering
\includegraphics[width=0.5\linewidth]{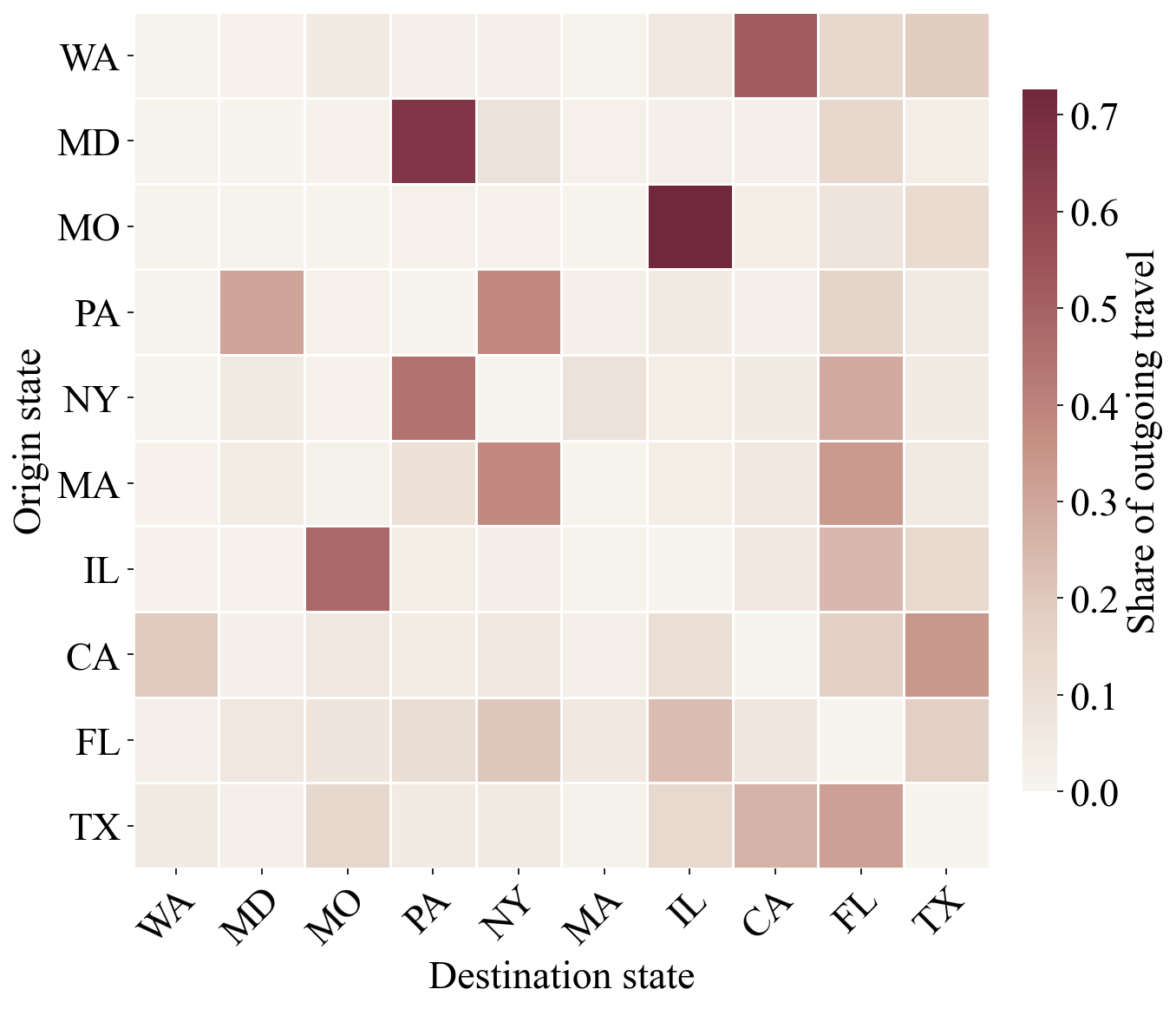}
    \caption{{Real interstate mobility graph.} Row-normalized Advan device-mobility flows among ten selected high-flow U.S.\ states in January 2021. Rows denote origins, columns denote destinations, and color indicates each destination's share of an origin's outgoing travel among the displayed states. The asymmetric, nonuniform matrix provides mobility-edge weights to the graph world model.}
    \label{fig:mobility}
\end{figure}

Figure~\ref{fig:mobility} shows a January 2021 snapshot of the row-normalized Advan mobility graph for ten selected high-flow U.S.\ states. Rows denote origins and columns denote destinations. The matrix exhibits directed, heterogeneous connectivity, with several dominant interstate links and long-range flows involving California, Florida, and Texas. EpiMind uses these flows as graph-edge weights, allowing the GF-RSSM to learn nonuniform neighbor contributions rather than assuming homogeneous regional mixing.

\subsection{Real-Context Prediction and Projection}
\label{app:realcontext_extended}

\begin{figure*}[t]
    \centering
    \begin{subfigure}[t]{0.45\textwidth}
        \caption*{{(a) Policy-conditioned rollout \ \ \ \ \ \ \ \ \ \ \ \ }}
        \includegraphics[width=\linewidth, height=5.2cm]{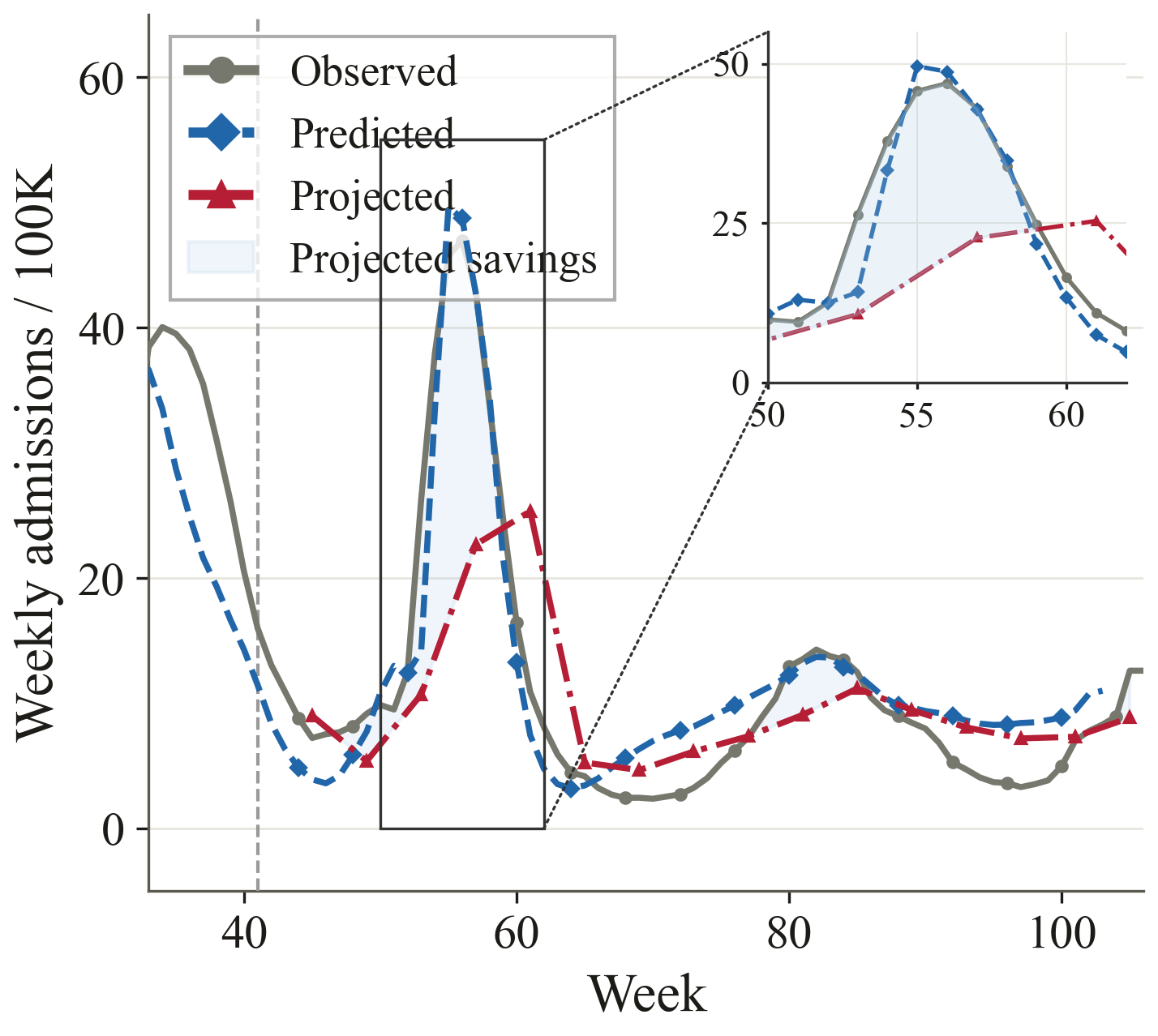}
        \label{fig:texas_trajectory}
    \end{subfigure}
    \hfill
    \begin{subfigure}[t]{0.5\textwidth}
        \caption*{{(b) Two-gate planner evaluation \ \ \ \ \ \ \ \ \ \ \ }}
        \includegraphics[width=\linewidth, height=5.2cm]{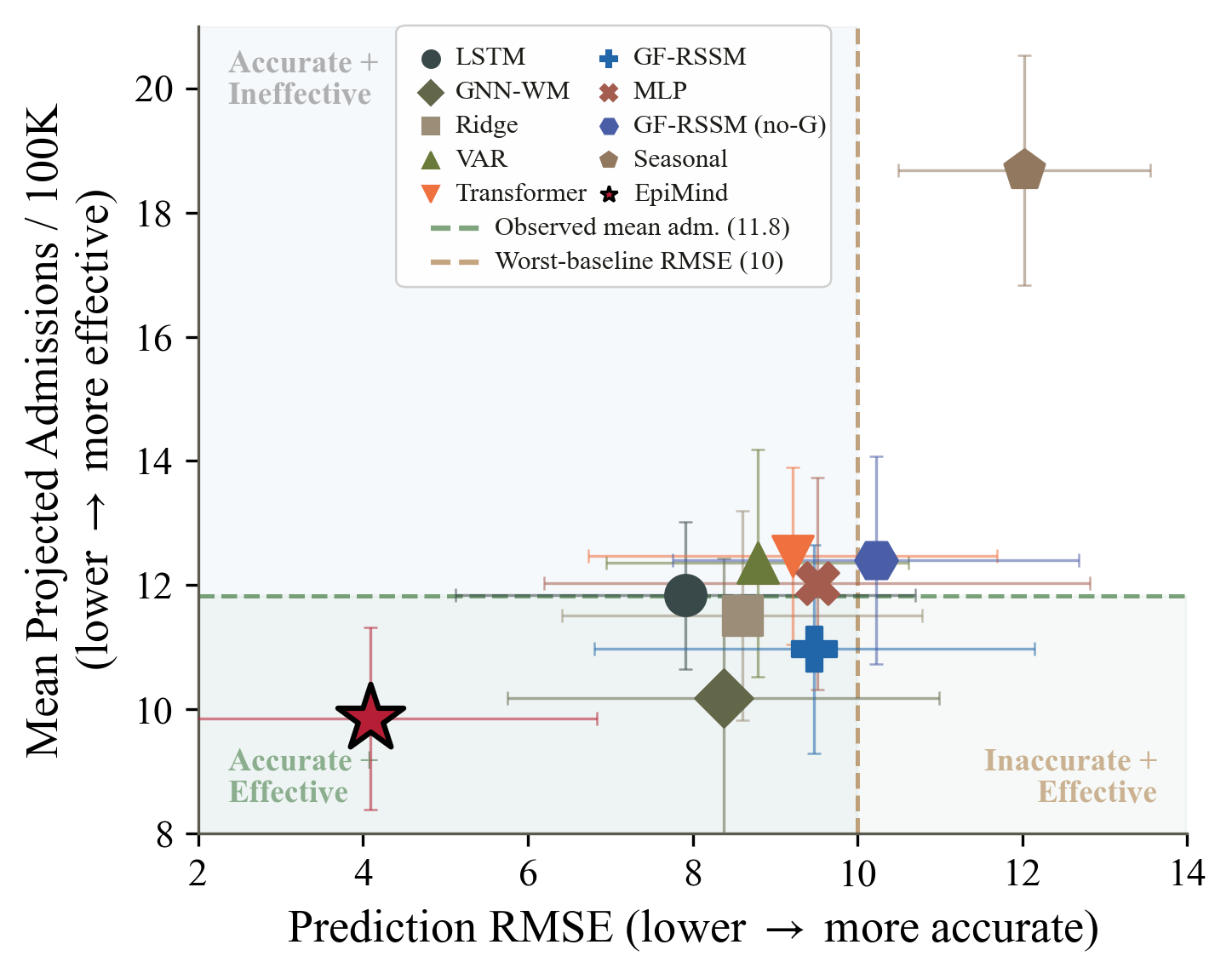}
        \label{fig:texas_scatter}
    \end{subfigure}
    \caption{
    {Texas real-context evaluation.}
    {(a)} Observed admissions, the held-out forecast under historical actions, and the model-relative rollout under EpiMind's proposed policy. The inset enlarges weeks 50--62; shading denotes the predicted difference between the historical- and proposed-policy rollouts. 
    {(b)} Forecast RMSE versus model-relative admissions projected under each model's optimized policy. Dashed lines mark the observed-policy admission mean and the selected forecast-error reference. Lower values are preferable on both axes.}
    \label{fig:texas_only}
\end{figure*}

Figure~\ref{fig:texas_only} examines whether forecast-capable models provide both accurate predictions and useful planning signals. \model{} attains the lowest forecast RMSE (4.1) and projects 9.9 weekly admissions per 100K, compared with 11.8 under the historical-policy reference. Several baselines also project admissions below this reference, but with substantially larger forecast errors. These results distinguish predictive fidelity from projected policy quality; because the proposed policies were not executed, the vertical axis represents model-relative outcomes rather than realized policy effects.

\begin{figure}[t] 
    \centering 
    \includegraphics[width=\linewidth] {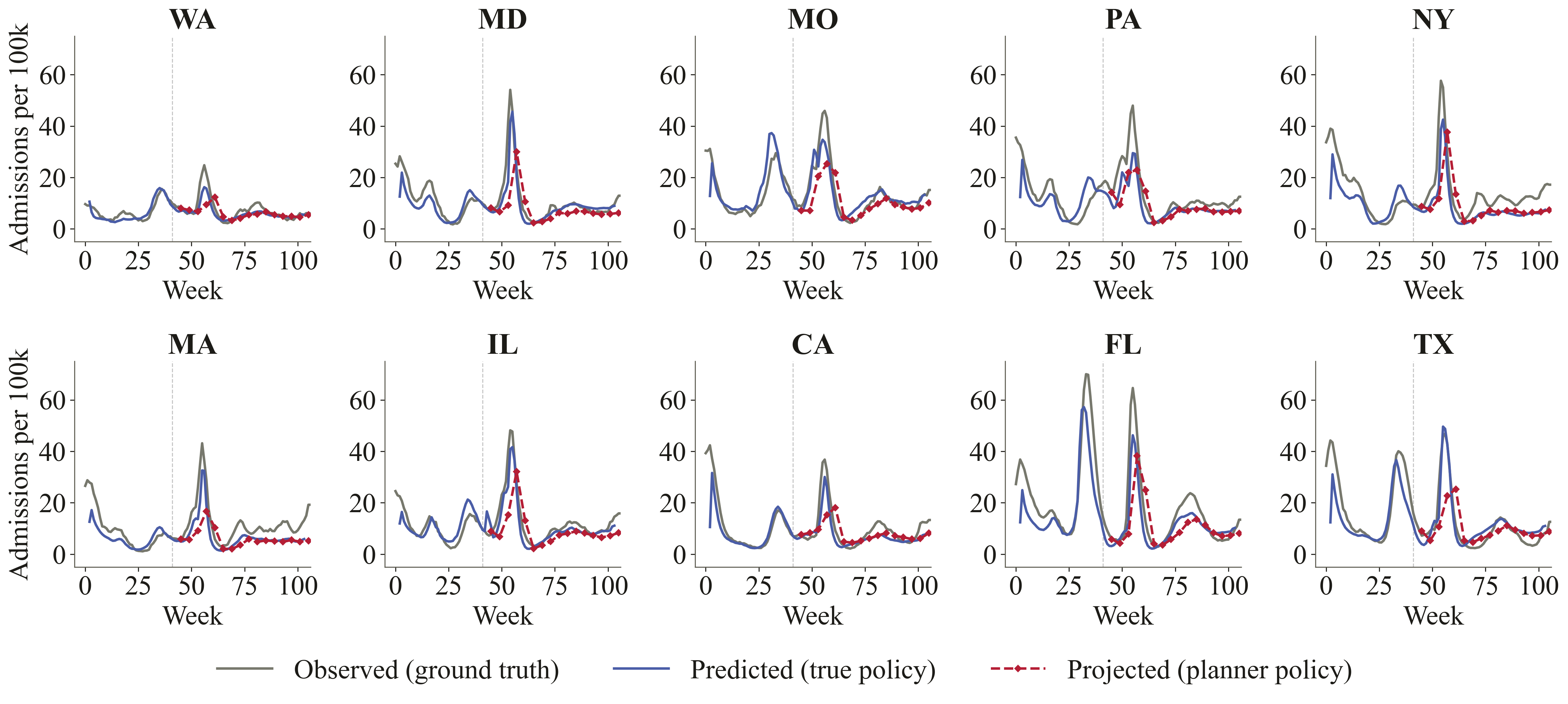} 
    \caption{
    {Multi-state forecasting and planner projections.} Observed weekly admissions (gray), forecasts under the recorded policy (blue), and model-relative rollouts under EpiMind's proposed policy (red) for ten mobility-connected U.S.\ states. The vertical dashed line marks the training cutoff. Planner projections represent unexecuted counterfactuals, not observed outcomes.} 
    \label{fig:multistate_rollouts} 
\end{figure}

Figure~\ref{fig:multistate_rollouts} illustrates EpiMind's shared-model behavior across heterogeneous regional trajectories. The historical-policy forecast tracks the timing of major admission waves, although peak magnitude is imperfectly calibrated in several states. The planner rollouts produce state-specific trajectories from the same mobility-coupled model. Because these policies were not executed, the projected reductions are interpreted as model-relative policy comparisons rather than causal effects.


\newpage

\end{document}